\documentclass[11pt,bezier,amstex]{article}

\usepackage{graphicx,verbatim}
\usepackage[rflt]{floatflt}
\usepackage{epsfig,times,subfigure,graphicx}
\usepackage{amsmath,amssymb,amsopn,algorithm,algorithmic,theorem,float,bbm,bm,enumerate,color,multirow}
\usepackage{fullpage,setspace}
\usepackage{rotating}
\usepackage{array}
\usepackage{hyperref}
\usepackage{url}
\usepackage{url}
\usepackage{chngcntr}
\usepackage{nameref}
\usepackage{zref-xr}
\zxrsetup{toltxlabel}

\newcommand{\beq}{\begin{equation}}
\newcommand{\eeq}{\end{equation}}

\newcommand{\htn}{\hat{\theta}_{\lambda_n}}
\newcommand{\hdn}{\hat{\Delta}_n}

\newcommand\I{\mathbb{I}}

\newcommand\s{\mathbb{S}}
\newcommand\R{\mathbb{R}}

\renewcommand{\u}{\mathbf{u}}

\newcommand{\y}{\mathbf{y}}

\newcommand{\cL}{{\cal L}}

\newcommand{\cN}{{\cal N}}

\newcommand{\cF}{{\cal F}}

\newcommand{\vertiii}[1]{{\left\vert\kern-0.25ex\left\vert\kern-0.25ex\left\vert #1
    \right\vert\kern-0.25ex\right\vert\kern-0.25ex\right\vert}}

\newcommand{\myref}[1]{(\ref{#1})}

\DeclareMathOperator{\argmin}{argmin}

\DeclareMathOperator{\cone}{cone}

\newcounter{exampleI}
{\theorembodyfont{\rmfamily} \theoremstyle{plain} }

\newcounter{exampleII}
{\theorembodyfont{\rmfamily} \theoremstyle{plain} }

\newcounter{exampleIII}
{\theorembodyfont{\rmfamily} \theoremstyle{plain} }

{\theorembodyfont{\rmfamily} }
{\theorembodyfont{\rmfamily} \newtheorem{defn}{Definition}}
\newtheorem{theo}{Theorem}

\newtheorem{lemm}{Lemma}

\newcommand{\proof}{\noindent{\itshape Proof:}\hspace*{1em}}

\newcommand{\qed}{\nolinebreak[1]~~~\hspace*{\fill} \rule{5pt}{5pt}\vspace*{\parskip}\vspace*{1ex}}

\newcommand {\commentout}[1] {}

\def\ints{{{\rm Z} \kern -.35em {\rm Z} }}  
\def\smallints{{{\rm Z} \kern -.3em {\rm Z} }}  
\def\pints{{{\rm I} \kern -.15em {\rm N} }}      
\newcommand{\reals}{\mathbb R}

\def\cplx{{{\rm I} \kern -.45em {\rm C} }}       
\def\l2{\rm {\mathcal L}^{2}(\reals)}            

\newtheorem{nad}{Notation and Definitions}[section]

\newcommand{\be}{\begin{eqnarray}}
\newcommand{\ee}{\end{eqnarray}}
\newcommand{\bea}{\begin{eqnarray}}
\newcommand{\eea}{\end{eqnarray}}
\newcommand{\beaa}{\begin{eqnarray*}}
\newcommand{\eeaa}{\end{eqnarray*}}
\newcommand{\bnad}{\begin{nad}}
\newcommand{\enad}{\end{nad}}

\title{Estimation with Norm Regularization}
\author{Arindam Banerjee \qquad  Sheng Chen \qquad Farideh Fazayeli \qquad Vidyashankar Sivakumar \vspace*{2mm}
\\
\{banerjee,shengc,farideh,sivakuma@cs.umn.edu\}\vspace*{2mm}\\
Department of Computer Science \& Engineering\\
University of Minnesota, Twin Cities}

\begin{document}

\maketitle

\begin{abstract}
Analysis of non-asymptotic estimation error and structured statistical recovery based on norm regularized regression, such as Lasso, needs to consider four aspects: the norm, the loss function, the design matrix, and the noise model. This paper presents generalizations of such estimation error analysis on all four aspects compared to the existing literature. We characterize the restricted error set where the estimation error vector lies, establish relations between error sets for the constrained and regularized problems, and present an estimation error bound applicable to {\em any} norm. Precise characterizations of the bound is presented for isotropic as well as anisotropic subGaussian design matrices, subGaussian noise models, and convex loss functions, including least squares and generalized linear models. Generic chaining and associated results play an important role in the analysis. A key result from the analysis is that the sample complexity of all such estimators depends on the Gaussian width of a spherical cap corresponding to the restricted error set. Further, once the number of samples $n$ crosses the required sample complexity, the estimation error decreases as $\frac{c}{\sqrt{n}}$, where $c$ depends on the Gaussian width of the unit norm ball.

\end{abstract}

\section{Introduction}
\label{sec:intro}
Over the past decade, progress has been made in developing non-asymptotic bounds on the estimation error of structured parameters based on norm regularized regression. Such estimators are usually of the form~\cite{tibs96,nrwy12,buva11}:
\beq
\hat{\theta}_{\lambda_n}~ = ~\underset{\theta \in \R^p}{\argmin} ~ \cL(\theta;Z^n) + \lambda_n R(\theta)~,
\label{eq:est1}
\eeq
where $R(\theta)$ is a suitable norm, $\cL(\cdot)$ is a suitable loss function, $Z^n = \{(y_i,X_i)\}_{i=1}^n$ where $y_i \in \R, X_i \in \R^p$ is the training set, and $\lambda_n >0$ is a regularization parameter.
The optimal parameter $\theta^*$ is often assumed to be `structured,'  usually characterized or approximated as a small value according to some norm $R(\cdot)$. Recent work has viewed such characterizations in terms of atomic norms, which give the tightest convex relaxation of a structured set of atoms in which $\theta^*$ belongs~\cite{crpw12}.
Since $\hat{\theta}_{\lambda_n}$ is an estimate of the optimal structure $\theta^*$, the focus has been on bounding a suitable measure of the error vector $\hdn=(\hat{\theta}_{\lambda_n} - \theta^*)$, e.g., the $L_2$ norm $\| \hdn \|_2$.

To understand the state-of-the-art on non-asymptotic bounds on the estimation error for norm-regularized regression, four aspects of \myref{eq:est1} need to be considered: (i) the norm $R(\theta)$, (ii) properties of the design matrix $X = [X_1 \cdots X_n]^T \in \R^{n \times p}$, (iii) the loss function $\cL(\cdot)$, and (iv) the noise model, typically in terms of $\omega_i = y_i - E[y|X_i]$. Most of the literature has focused on a linear model: $y = X \theta + \omega$, and a squared-loss function:
$\cL(\theta;Z^n)  = \frac{1}{n} \| y - X \theta \|_2^2 = \frac{1}{n} \sum_{i=1}^n (y_i - \langle \theta, X_i \rangle)^2$.
Early work on such estimators focussed on the $L_1$ norm~\cite{zhyu06,wain09,meyu09}, and led to sufficient conditions on the design matrix $X$, including the restricted-isometry properties (RIP)~\cite{cata05,cart06} and restricted eigenvalue (RE) conditions~\cite{birt09,nrwy12,rawy10}. While much of the development has focussed on isotropic Gaussian design matrices, recent work has extended the analysis for $L_1$ norm to correlated Gaussian designs~\cite{rawy10} as well as anisotropic sub-Gaussian design matrices~\cite{ruzh13}.

Building on such development,~\cite{nrwy12} presents a unified framework for the case of decomposable norms and also considers generalized linear models (GLMs) for certain norms such as $L_1$. Two key insights are offered in~\cite{nrwy12}: first, the error vector $\hdn$ lies in a restricted set, a cone or a star, for suitably large $\lambda_n$, and second, the loss function needs to satisfy restricted strong convexity (RSC), a generalization of the RE condition, on the restricted error set for the analysis to work out.


For isotropic Gaussian design matrices, additional progress has been made. \cite{crpw12} considers a constrained
estimation formulation for all atomic norms, where the gain condition, equivalent to the RE condition, uses Gordon's inequality~\cite{gord85,gord88,ledo13} and is succinctly represented in terms of the Gaussian
width of the intersection of the cone of the error set and a unit ball/sphere. \cite{oyth13} considers three related formulations for generalized Lasso problems, establish recovery guarantees based on Gordon's inequality, and quantities related to the Gaussian width. Sharper analysis for recovery has been considered in~\cite{almt14}, yielding a precise characterization of phase transition behavior using quantities related to the Gaussian width. \cite{plve13} consider a linear programming estimator in a 1-bit compressed sensing setting and, interestingly, the concept of Gaussian width shows up in the analysis. In spite of the advances, with a few notable exceptions~\cite{trop14,vers14}, most existing results are restricted to isotropic Gaussian design matrices. Further, while a suitable scale for $\lambda_n$ is known for special cases such as the $L_1$, a general analysis applicable to any norm $R(\cdot)$ has not been explored in the literature.


In this paper, we consider structured estimation problems with norm regularization of the form~\myref{eq:est1}, and present a unified analysis which substantially generalizes existing results on all four pertinent aspects: the norm, the design matrix, the loss,
and the noise model. The analysis we present applies to {\em all} norms, and the results can be divided into three groups: characterization of the error set and recovery guarantees, characterization of the regularization parameter $\lambda_n$, and characterization of the restricted eigenvalue conditions or restricted strong convexity. We provide a summary of the key results below.

{\bf Restricted error set:} We start with a characterization of the error set $E_r$ in which the error vector $\hdn$ belongs. For a suitably large $\lambda_n$, we show that $\hdn$ belongs to the restricted error set
\beq
E_r = \left\{ \Delta \in \R^p ~ \left| ~ R(\theta^* + \Delta) \leq R(\theta^*) + \frac{1}{\beta} R(\Delta) \right. \right\}~,
\eeq
where $\beta > 1$ is a constant. The restricted error set has interesting structure, and forms the basis of subsequent analysis for bounds on $\| \hdn \|_2$.

{\bf Regularized vs.~constrained estimators:} As an alternative to regularized estimators, the literature has considered constrained estimators which directly focus on minimizing $R(\theta)$ under suitable constraints determined by the noise $(y-X\theta)$ and/or the design matrix $X$~\cite{cata05,birt09,crpw12,chcb14}. A recent example of such a constrained estimator is the generalized Dantzig selector (GDS)~\cite{chcb14}, which generalizes the Dantzig selector~\cite{cata07} corresponding to the $L_1$ norm, and is given by:
\beq
\hat{\theta}_{\gamma_n} ~= ~\underset{\theta \in \R^p}{\argmin}~R(\theta) \quad \text{s.t.} \quad R^*( X^T (y - X\theta^*)) \leq \gamma_n~,
\label{eq:est22}
\eeq
where $R^*(\cdot)$ denotes the dual norm of $R(\cdot)$. One can show~\cite{crpw12,chcb14} that the restricted error set for such constrained estimators are of the form:
\beq
E_c = \left\{ \Delta \in \R^p ~ \left| ~ R(\theta^* + \Delta) \leq R(\theta^*) \right. \right\}~.
\eeq
One can readily see that $E_r$ is larger than $E_c$, i.e., $E_c \subseteq E_r$, and $E_r$ approaches $E_c$ as $\beta$ increases. We establish a geometric relationship between the two sets, which will possibly help in transforming analysis done on regularized estimators as in \myref{eq:est1} to corresponding constrained estimators as in \myref{eq:est22} and vice versa. Let $B_2^p$ be a $L_2$ ball of radius 1 in $\R^p$. Then, with $A_r = E_r \cap B_2^p, A_c = E_c \cap B_2^p$, and $\bar{A}_c = \cone(E_c) \cap B_2^p$, assuming $\| \theta^* \|_2=1, \beta=2$, we show that
\beq
w(A_c) \leq w(A_r) \leq 3 w(\bar{A}_c) ~,
\eeq
where $w(A) = E_g[\sup_{a \in A} \langle a, g \rangle]$, with $g \sim N(0,\I_{p \times p})$ being an isotropic Gaussian vector, denotes the Gaussian width\footnote{A gentle exposition to Gaussian width and some of its properties is given in Appendix~\ref{app:back}.} of the set $A$~\cite{ledo13,bame02,crpw12,tala05}. Note that $A_r$ corresponds to the spherical cap of the error set $E_r$, and $A_c$ corresponds to the spherical cap of the error cone $\cone(E_c)$ at the unit ball. Interestingly, the above relationship between the widths of these spherical caps is geometric, and applies for any norm $R(\cdot)$.

We establish a more general version of the above relationship. Let $\rho B_2^p$ denote a $L_2$ ball of any radius $\rho$ in $\R^p$. Then, with $A_r^{(\rho)} = E_r \cap \rho B_2^p, A_c^{(\rho)}= E_c \cap \rho B_2^p$, and $\bar{A}_c^{(\rho)} = \cone(E_c) \cap \rho B_2^p$, we show that
\beq
w(A_c^{(\rho)}) \leq w(A_r^{(\rho)}) \leq \left ( 1 + \frac{2}{\beta - 1} \frac{\|\theta^*\|_2}{\rho} \right ) w(\bar{A}_c^{(\rho)})~.
\eeq
As before, except for the scaling constants, the relationship between the restricted error sets is geometric, and does not change based on the choice of the norm $R(\cdot)$.

For the special case of $L_1$ norm, \cite{birt09} considered a simultaneous analysis of the Lasso and the Dantzig selector, and characterized the structure of the error sets for regularized and constrained sets for the special case of $L_1$ norm. Further, while the characterization in \cite{birt09} was also geometric, it was not based on Gaussian widths. In contrast, our results apply to any norm, not just $L_1$, and the geometric characterization is based on Gaussian widths. The utility of the Gaussian width based characterization becomes evident later when we establish sample complexity results for Gaussian and sub-Gaussian random matrices in terms of Gaussian widths of spherical caps.

{\bf Bounds on estimation error:} We establish bounds on the estimation error $\hdn$ under two assumptions, which are subsequently shown to hold with high probability for sub-Gaussian designs and noise models. The first assumption is that the regularization parameter $\lambda_n$ is suitably large. In particular, for any $\beta > 1$, the regularization parameter $\lambda_n$ needs to satisfy
\beq
\lambda_n \geq \beta R^*(\nabla \cL(\theta^*;Z^n))~,
\label{eq:assum1}
\eeq
where $R^*(\cdot)$ denotes the dual norm of $R(\cdot)$.
The second assumption is that the design matrix $X \in \R^{n \times p}$ satisfies the restricted strong convexity (RSC) condition~\cite{birt09,nrwy12} in the error set $E_r$, in particular, there exists a suitable constant $\kappa > 0$ so that
\beq
\delta \cL(\Delta,\theta^*) \triangleq \cL(\theta^*+ \Delta) - \cL(\theta^*) - \langle \nabla \cL(\theta^*),\Delta \rangle \geq \kappa \| \Delta \|_2^2~ \quad \forall \Delta \in E_r~.
\label{eq:assum2}
\eeq
With such suitably large $\lambda_n$ and $\cL$ satisfying the RSC condition, we establish the following bound:
\beq
\| \hdn \|_2 \leq c \psi(E_r) \frac{\lambda_n}{\kappa}~,
\label{eq:result1}
\eeq
where $\psi(E_r) = \sup_{u \in E_r} \frac{R(u)}{\| u\|_2}$ is a norm compatibility constant~\cite{nrwy12}, and $c>0$ is a constant. Note that the above bound is deterministic, but relies on assumptions on $\lambda_n$ and $\kappa$. So, we focus on characterizations of $\lambda_n$ and $\kappa$ which hold with high probability for sub-Gaussian design matrices $X$ and sub-Gaussian noise $\omega$. Recent work in \cite{sibr15} has extended the analyses for sub-exponential distributions.



{\bf Bounds on the regularization parameter $\lambda_n$:} From \myref{eq:assum1} above, for the analysis to work, one needs to have $\lambda_n \geq \beta R^*(\nabla \cL(\theta^*;Z^n))$. There are a few challenges in getting a suitable bound for $\lambda_n$. First, the bound depends on $\theta^*$, but $\theta^*$ is unknown and is the quantity one is interested in estimating. Second, the bound depends on $Z^n$, the samples, and is hence random. The goal will be to bound the expectation $E[R^*(\nabla \cL(\theta^*;Z^n))]$ over all samples of size $n$, and obtain high-probability deviation bounds. Third, since the bound relies on the (dual) norm $R^*(\cdot)$ of a $p$-dimensional random vector, without proper care, the lower bound on $\lambda_n$ may end up having a large scaling dependency, say $\sqrt{p}$, on the ambient dimensionality. Since the error bound in \myref{eq:result1} is directly proportional to $\lambda_n$, such dependencies will lead to weak bounds.

In Section~\ref{sec:lambda}, we characterize the expectation $E[R^*(\nabla \cL(\theta^*;Z^n))]$ in terms of the geometry of the unit norm-ball of $R$, which leads to a sharp bound. Let $\Omega_R = \{ u \in \R^p | R(u) \leq 1\}$ denote the unit norm-ball. Then, for sub-Gaussian design matrices and squared loss, we show that
\beq
E[R^*(\nabla \cL(\theta^*;Z^n))] \leq \frac{c}{\sqrt{n}} w (\Omega_R)~,
\eeq
which scales as the Gaussian width of $\Omega_R$. Interestingly, for sub-Gaussian designs, one obtains the results in terms of the `sub-Gaussian width' of the unit norm-ball, which can be upper bounded by a constant times the Gaussian width using generic chaining~\cite{tala05}. The result can be extended to the case of anisotropic sub-Gaussian designs, where the constant $c$ starts depending on the maximum eigenvalue (operator norm) of the corresponding covariance matrix. Further, one can get high-probability versions of these bounds using related advances in generic chaining~\cite{tala05,tala14}. The results can also be extended to general convex losses, such as those from generalized linear models.


The above characterization allows one to choose $\lambda_n \geq \frac{c}{\sqrt{n}} w (\Omega_R)$. For the special case of $L_1$ regularization, $\Omega_R$ is the unit $L_1$ norm ball, and the corresponding Gaussian width $w(\Omega_R) \leq c_1 \sqrt{\log p}$, which explains the $\sqrt{\log p}$ term one finds in existing bounds for Lasso~\cite{nrwy12,buva11}. When working with other norms, one simply needs to get an upper bound on the corresponding $w(\Omega_R)$.


{\bf Restricted eigenvalue conditions:} When the loss function under consideration is the squared loss, the RSC condition in \myref{eq:assum2} reduces to the restricted eigenvalue (RE) condition on the design matrix.
Our analysis focuses on establishing the RE condition on $\bar{A}_r = \cone(E_r) \cap S^{p-1}$, the spherical cap obtained by intersecting the cone of the error set with the unit hypersphere, since it implies the RE condition on $E_r$.
For isotropic sub-Gaussian design matrices, a stronger two-sided restricted isometry property (RIP) holds, i.e., with high probability, for any $A \subseteq S^{p-1}$, we have
\beq
1 - c \frac{w(A)}{\sqrt{n}} \leq \inf_{u \in A} \frac{1}{n} \| X u \|^2 \leq \sup_{u \in A} \frac{1}{n} \| Xu\|^2 \leq 1 + c \frac{w(A)}{\sqrt{n}}
\eeq
where $w(A)$ is the Gaussian width of $A$. Thus, for say $n_0 = 4 c^2 w^2(\bar{A}_r)$, and for $n > n_0$, an RE condition of the form
\beq
\inf_{ u \in \bar{A}_r} \frac{1}{n} \| X u \|^2_2 ~~ \geq ~~ 1/2 ~,
\eeq
is satisfied with high probability. Instead of the constant to be $1/2$, one can have any constant less than 1, with suitable increase in $n_0$. Thus, one does not need to treat the RE condition as an assumption for isotropic sub-Gaussian designs---it always holds with high probability, with the phase transition happening at $O(w^2(\bar{A}_r))$ samples. The RIP results can be generalized to anisotropic sub-Gaussian designs, where additional constants depending on the restricted eigenvalues of the anisotropic covariance matrix $\Sigma$ show up, but the form of the bound stays similar. Our analysis techniques for the RIP results are based on generic chaining~\cite{tala05,tala14}, in particular a specific form developed in~\cite{klme05,mept07}.

{\bf Generalized linear models and restricted strong convexity:} For convex loss functions, such as those coming from generalized linear models (GLMs), the sample complexity and associated phase transition behavior is determined by the Restricted Strong Convexity (RSC) condition~\cite{nrwy12}. By generalizing our argument for RE conditions corresponding to square loss, we show that the RSC conditions are going to be satisfied for convex losses for subGaussian designs at the same order of sample complexity as that for squared loss. In particular, for  we show a high probability lower bound of the form
\beq
\inf_{u \in A}~ \delta \cL(u,\theta^*) \geq c_1  - c_2 \frac{w(A)}{\sqrt{n}}~,
\eeq
where the constants $c_1, c_2 > 0$ depend on the tail probabilities of the design matrix distribution. Specializing the result to $\bar{A}_r = \cone(E_r) \cap S^{p-1}$, we note that the sample complexity still scales as $O(w^2(\bar{A}_r))$, similar to the case of  squared loss. The result is thus a considerable generalization of earlier results on convex losses, such as GLMs, which had looked at specific norms and associated cones and/or did not express the results in terms of the Gaussian width of $A$~\cite{nrwy12}.



{\bf Putting everything together:} With the above results in place, from~\myref{eq:result1}, the main bound takes the form
\beq
\| \hdn \|_2  \leq c \frac{ \psi(E_r)}{\left[ c_1 - c_2 \frac{w(\bar{A}_r)}{\sqrt{n}} \right]_{+}} \frac{ w(\Omega_R)}{\sqrt{n}}
\eeq
with high probability, where $w(\Omega_R)$ is the Gaussian width of the unit norm ball, $w(\bar{A}_r)$ is the Gaussian width of the spherical cap corresponding to the error set $\cone(E_r)$, and the result is valid only when $n > n_0 = O(w^2(\bar{A}_r))$ which corresponds to the sample complexity. For the special case of $L_1$ norm, i.e., Lasso, the sample complexity $n_0$ is of the order $w^2(\bar{A}_r) = O(s\log p)$. Further, $w(\Omega_R) = \sqrt{\log p}$ and $\psi(E_r) = \sqrt{s}$. Plugging in these values, choosing $\beta = 2$, for $n > c_3 s \log p$, the bound $\| \hdn \|_2 \leq c \sqrt{\frac{s \log p}{n}}$ holds with probability. For other norms, one can simply plug-in the widths to get the corresponding sample complexity and non-asymptotic error bounds.


The rest of the paper is organized as follows: Section~\ref{sec:error} presents results on the restricted error set and deterministic error bounds under suitable bounds on the regularization parameter $\lambda_n$ and RSC assumptions. Section~\ref{sec:lambda} presents a characterization of $\lambda_n$ in terms of the Gaussian width of the unit norm ball for Gaussian as well as sub-Gaussian designs and noise. Section~\ref{sec:re} proves RE conditions and associated sample complexity results corresponding to squared loss functions. Results are presented for subGaussian designs, including anisotropic and correlated cases, and always in terms of the Gaussian width of the spherical cap corresponding to the error set. Section~\ref{sec:glm} presents RSC conditions corresponding to general convex losses arising from generalized linear models, and the results are again in terms of the Gaussian width of the spherical cap corresponding to the error set. We conclude in Section~\ref{sec:conc}.
All technical arguments and proofs are in the appendix, along with a gentle exposition to Gaussian widths and related results.

A brief word on the notation used. We denote random matrices as $X$, and random vectors as $X_i$ where $i$ may be an index to a row or column of a random matrix. Vector norms are denoted as $\| \cdot \|$, e.g., $\| X_i \|_2$ for a (random) vector $X_i$, and norms of random variables are denoted as $\vertiii{\cdot}$, e.g., $\vertiii{X}_2 = E[\| X \|_2]$.


%
%

\section{Restricted Error Set and Recovery Guarantees}
\label{sec:error}

In this section, we give a characterization of the restricted error set $E_r$ in which the error vector $\hdn = (\htn - \theta^*)$ lies, establish clear relationships between the error sets for the regularized and constrained problems, and finally establish upper bounds on the estimation error. The error bound is deterministic, but has quantities which involve $\theta^*, X, \omega$, for which we develop high probability bounds in Sections~\ref{sec:lambda}, \ref{sec:re}, and \ref{sec:glm}.

\begin{figure}
\centering
\includegraphics[trim = 3cm 1.6cm 3cm 3cm, clip, width=0.7\textwidth]{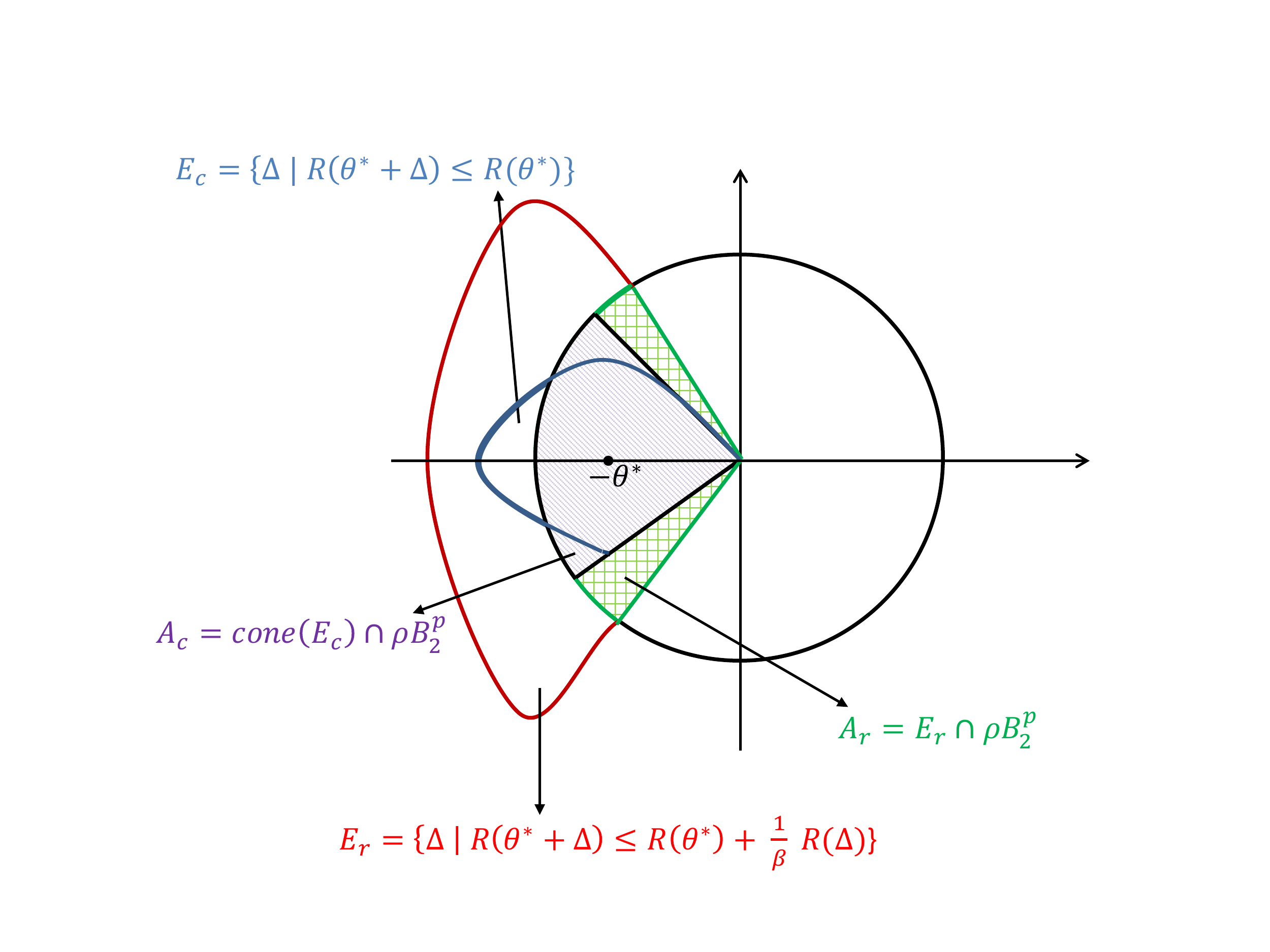}
\caption{Relationship between error set of the regularized problem ($A_r$, green region) and the constrained problem ($A_c$, gray region) after intersection with a ball of radius $\rho$. While $A_r$ will be larger in general, it will be within a constant factor of $A_c$ in terms of Gaussian width (best viewed in color).}
\label{fig:errSetRegCons}
\end{figure}

\subsection{The Restricted Error Set and the Error Cone}
We start with a characterization of the restricted error set $E_r$  where $\hdn$ will belong.
\begin{lemm}
For any $\beta > 1$, assuming
\beq
\lambda_n \geq \beta R^*(\nabla \cL(\theta^*;Z^n))~,
\label{eq:cond1}
\eeq
where $R^*(\cdot)$ is the dual norm of $R(\cdot)$. Then the error vector $\hat{\Delta}_n = \hat{\theta}_{\lambda_n} - \theta^*$ belongs to the set
\beq
E_r = E_r(\theta^*,\beta) = \left\{ \Delta \in \R^p ~ \left| ~ R(\theta^* + \Delta) \leq R(\theta^*) + \frac{1}{\beta} R(\Delta) \right. \right\}~.
\label{eq:error1}
\eeq
\vspace*{-3mm}
\label{lem:reserrset}
\end{lemm}
The restricted error set $E_r$ need not be convex for general norms. Interestingly, for $\beta=1$, the inequality in \myref{eq:error1} is just the triangle inequality, and is satisfied by all $\Delta$. Note that $\beta > 1$ restricts the set of $\Delta$ which satisfy the inequality, yielding the restricted error set.
In particular, $\Delta$ cannot go in the direction of $\theta^*$, i.e., $\Delta \neq \alpha \theta^*$ for any $\alpha > 0$.
Further, note that the condition in~\myref{eq:cond1} is similar to that in~\cite{nrwy12} for $\beta=2$, but the above characterization holds for any norm, not just decomposable norms~\cite{nrwy12}.


While $E_r$ need not be a convex set, we establish a relationship between $E_r$ and the error set $E_c$ corresponding to constrained estimators~\cite{cata05,birt09,crpw12,chcb14}. A recent example of such a constrained estimator is the generalized Dantzig selector (GDS)~\cite{chcb14} given by:
\beq
\hat{\theta}_{\gamma_n} ~= ~\underset{\theta \in \R^p}{\argmin}~R(\theta) \quad \text{s.t.} \quad R^*( X^T (y - X\theta^*)) \leq \gamma_n~,
\label{eq:est2}
\eeq
where $R^*(\cdot)$ denotes the dual norm of $R(\cdot)$. One can show~\cite{crpw12,chcb14} that the restricted error set for such constrained estimators~\cite{crpw12,chcb14,trop14} are of the form:
\beq
E_c = \left\{ \Delta \in \R^p ~ \left| ~ R(\theta^* + \Delta) \leq R(\theta^*) \right. \right\}~.
\eeq
By definition, it is easy to see that $E_c$ is always convex, and that $E_c \subseteq E_r$, as shown schematically in Figure~\ref{fig:errSetRegCons}.

%
The following results establishes a relationship between $E_r$ and $E_c$ in terms of their Gaussian widths.
\begin{theo}
Let $A_c^{(\rho)} = E_c \cap \rho B_2^p$, $A_r^{(\rho)} = E_r \cap \rho B_2^p$, and $\bar{A}_c^{(\rho)} = C_c \cap \rho B_2^p$, where $\rho B_2^p = \{ u | \|u\|_2 \leq \rho\}$ is the $L_2$ ball of any radius $\rho > 0$. Then, for any $\beta > 1$ we have
\beq
w(A_c^{(\rho)}) \leq w(A_r^{(\rho)}) \leq \left ( 1 + \frac{2}{\beta - 1} \frac{\|\theta^*\|_2}{\rho} \right ) w(\bar{A}_c^{(\rho)}) ~,
\eeq
where $w(A)$ denotes the Gaussian width of any set $A$ given by: $w(A) = E_g\left[\underset{a\in A}{\sup}~\langle a,g \rangle\right]$, where $g$ is an isotropic Gaussian random vector, i.e., $g \sim N(0, \mathbb{I}_{p \times p})$.
\label{thm:relconsreg}
\end{theo}
Thus, the Gaussian width of the error sets of regularized and constrained problems are closely related. See Figure \ref{fig:errSetRegCons} for more details.
In particular, for $\| \theta^* \|_2 = 1$, with $\rho = 1, \beta = 2$, we have $w(A_c) \leq w(A_r) \leq 3 w(\bar{A}_c)$ as introduced in Section \ref{sec:intro}.
Related observations have been made for the special case of the $L_1$ norm~\cite{birt09}, although past work did not provide an explicit characterization in terms of Gaussian widths. The result also suggests that it is possible to move between the error analysis of the regularized and the constrained versions of the estimation problem.
%
%
%
%
%
%

\begin{figure}
\centering
\subfigure[Loss and regularizer.]{\includegraphics[width=0.4\textwidth]{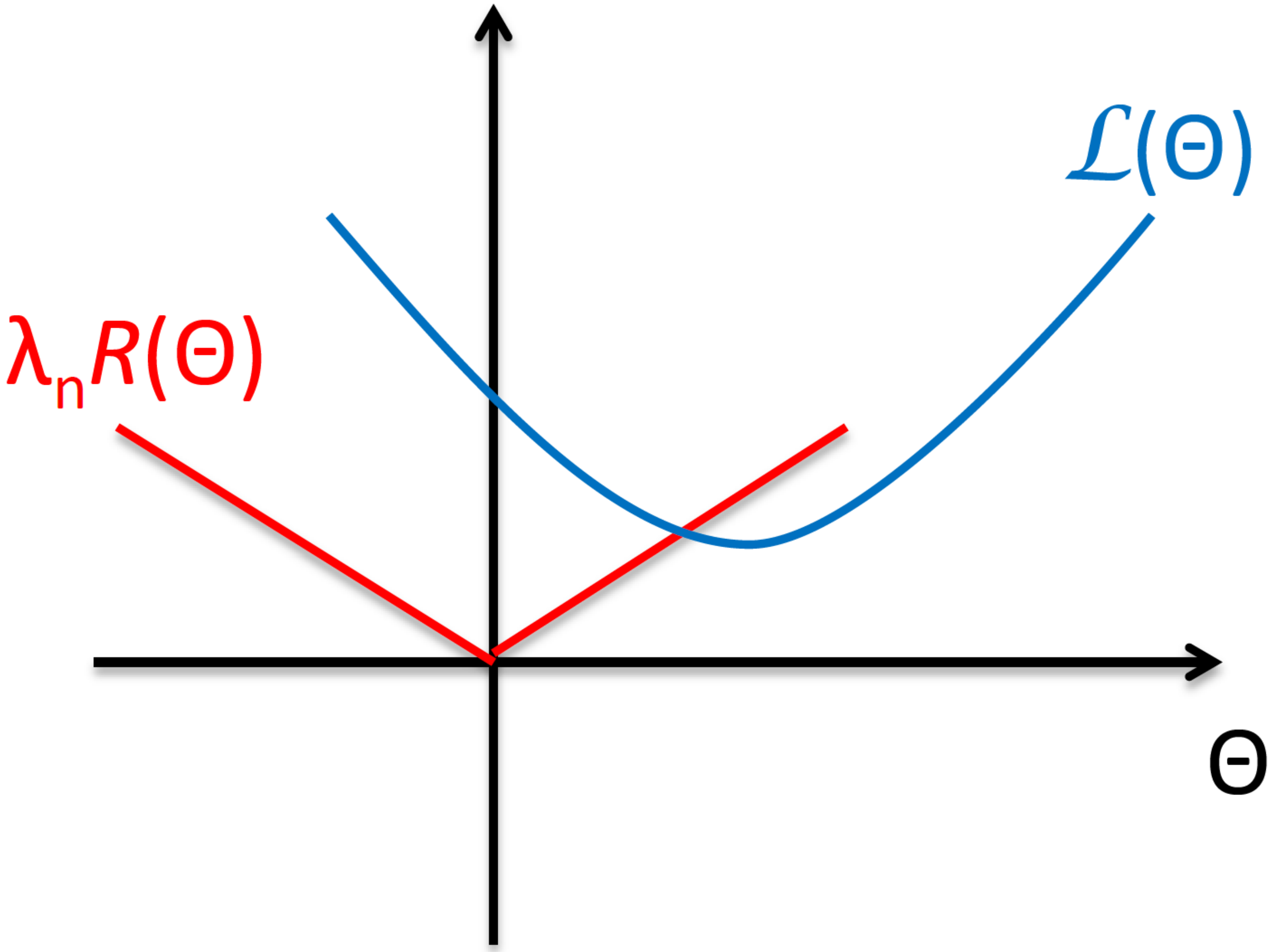}}
\hspace*{8mm}
\subfigure[Overall objective.]{\includegraphics[width=0.5\textwidth]{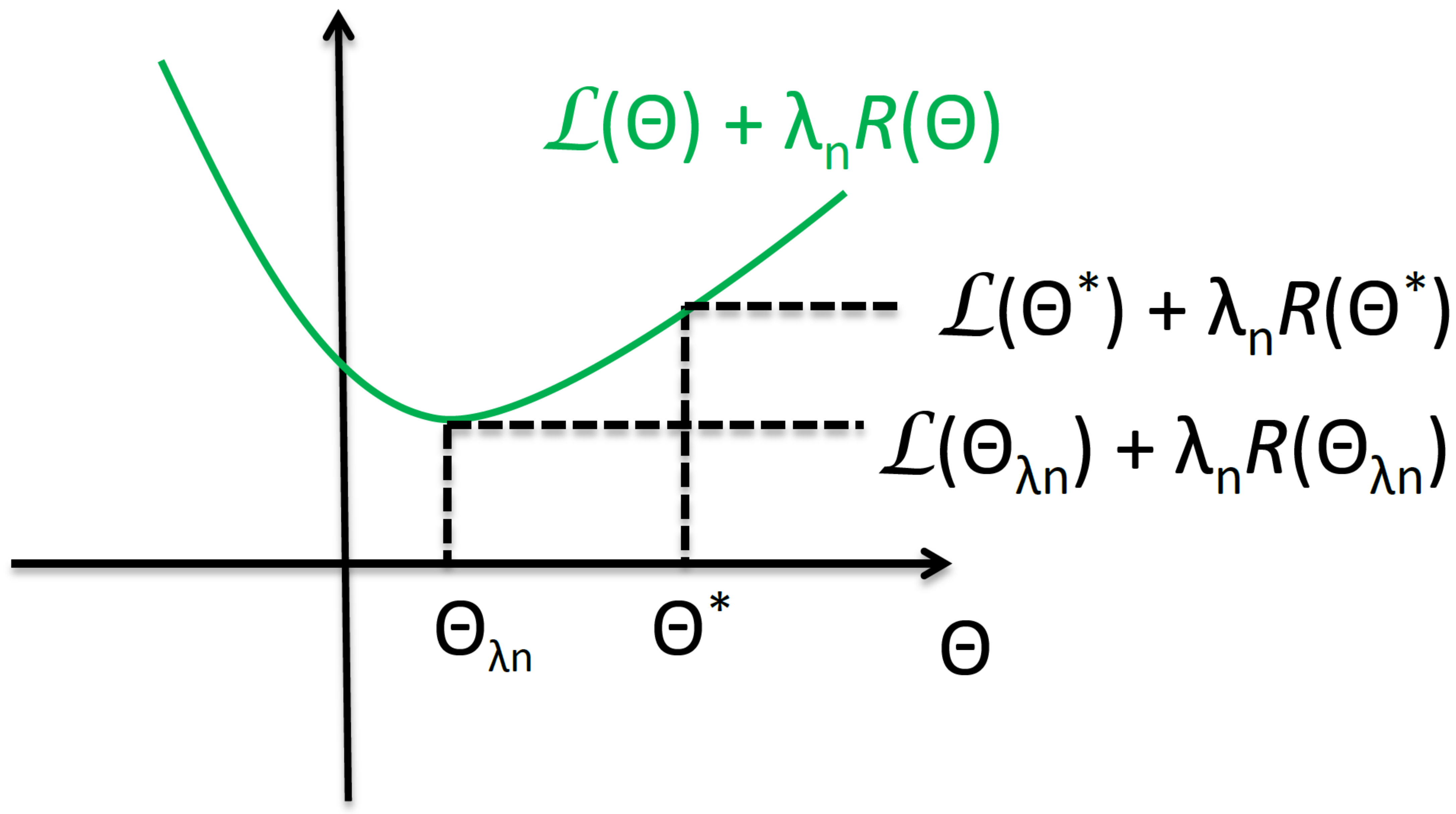}}
\caption{Schematic for norm regularized objective functions considered. The finite sample estimate $\hat{\theta}_{\lambda_n}$ has lower empirical loss than the optimum $\theta^*$. Bounding the difference between the losses yields a bound on $\| \hat{\theta}_{\lambda_n} - \theta^*\|$.}
\label{fig:rsc}
\end{figure}

\subsection{Recovery Guarantees}
In order to establish recovery guarantees, we start by assuming that restricted strong convexity (RSC) is satisfied by the loss function in $E_r$, the error set, so that for any $\Delta \in E_r$, there exists a suitable constant $\kappa$ so that
\beq
\delta \cL (\Delta, \theta^*) \triangleq \cL(\theta^*+\Delta) - \cL(\theta^*) - \langle \nabla \cL(\theta^*),\Delta \rangle \geq \kappa \| \Delta \|_2^2~.
\label{eq:rsc}
\eeq
In Sections~\ref{sec:re} and \ref{sec:glm}, we establish precise forms of the RSC condition for a wide variety of design matrices and loss functions. In order to establish recovery guarantees, we focus on the quantity
\beq
\cF(\Delta) = \cL(\theta^*+\Delta) - \cL(\theta^*) + \lambda_n (R(\theta^* + \Delta) - R(\theta^*))~.
\eeq
Since $\htn = \theta^* + \hdn$ is the estimated parameter, i.e., $\htn$ is the minimum of the objective, we clearly have $\cF(\hdn) \leq 0$, which implies a bound on $\| \hdn \|_2$. Unlike previous analysis, the bound can be established without making any additional assumptions on the norm $R(\theta)$.
We start with the following result, which expresses the upper bound on $\| \hdn \|_2$ in terms of the gradient of the objective at $\theta^*$.
\begin{lemm}
Assume that the RSC condition is satisfied in $E_r$ by the loss $\cL(\cdot)$ with parameter $\kappa$. With $\hdn = \htn - \theta^*$, for any norm $R(\cdot)$, we have
\beq
\| \hdn \|_2 \leq \frac{1}{\kappa} \| \nabla \cL(\theta^*) + \lambda_n \nabla R(\theta^*) \|_2~,
\eeq
where $\nabla R(\cdot)$ is any sub-gradient of the norm $R(\cdot)$.
\label{lem:recbnderror}
\end{lemm}
Figure \ref{fig:rsc} illustrates the above results.
Note that the right hand side is simply the $L_2$ norm of the gradient of the objective evaluated at $\theta^*$. For the special case when $\htn = \theta^*$, the gradient of the objective is zero, implying correctly that $\| \hdn \|_2 = 0$. While the above result provides useful insights about the bound on $\| \hdn \|_2$, the quantities on the right hand side depend on $\theta^*$, which is unknown. We present another form of the result in terms of quantities such as $\lambda_n$, $\kappa$, and the norm compatibility constant $\psi(E_r) = \sup_{\u \in E_r} \frac{R(\u)}{\| \u \|_2}$, which are often easier to compute or bound.
\begin{theo}
Assume that the RSC condition is satisfied in $E_r$ by the loss $\cL(\cdot)$ with parameter $\kappa$. With $\hdn = \htn - \theta^*$, for any norm $R(\cdot)$, we have
\beq
\| \hdn \|_2 \leq  \psi(E_r) \frac{1+\beta}{\beta} \frac{\lambda_n}{\kappa} ~.
\label{eq:recover}
\vspace*{-5mm}
\eeq
\label{thm:recover}
\end{theo}
The above result is deterministic, but contains $\lambda_n$ and $\kappa$. In Section~\ref{sec:lambda}, we give precise characterizations of $\lambda_n$, which needs to satisfy~\myref{eq:cond1}. In Sections~\ref{sec:re} and \ref{sec:glm}, we characterize the RSC condition constant $\kappa$ for different losses and a variety of design matrices.

\begin{figure}
\centering
\subfigure[Gradients of loss and norm.]{\includegraphics[width=0.44\textwidth]{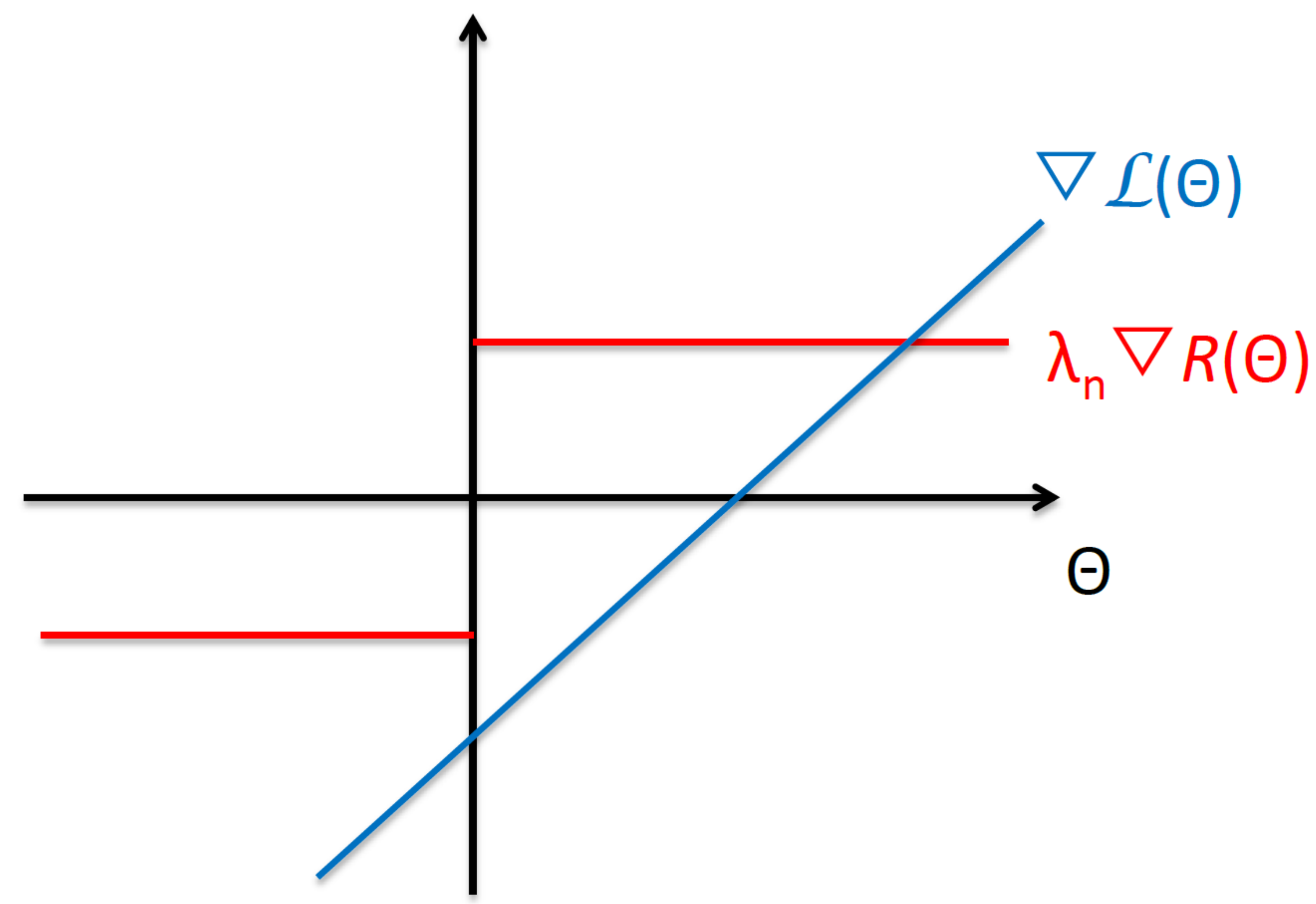}}
\hspace*{8mm}
\subfigure[Gradient of the overall objective]{\includegraphics[width=0.5\textwidth]{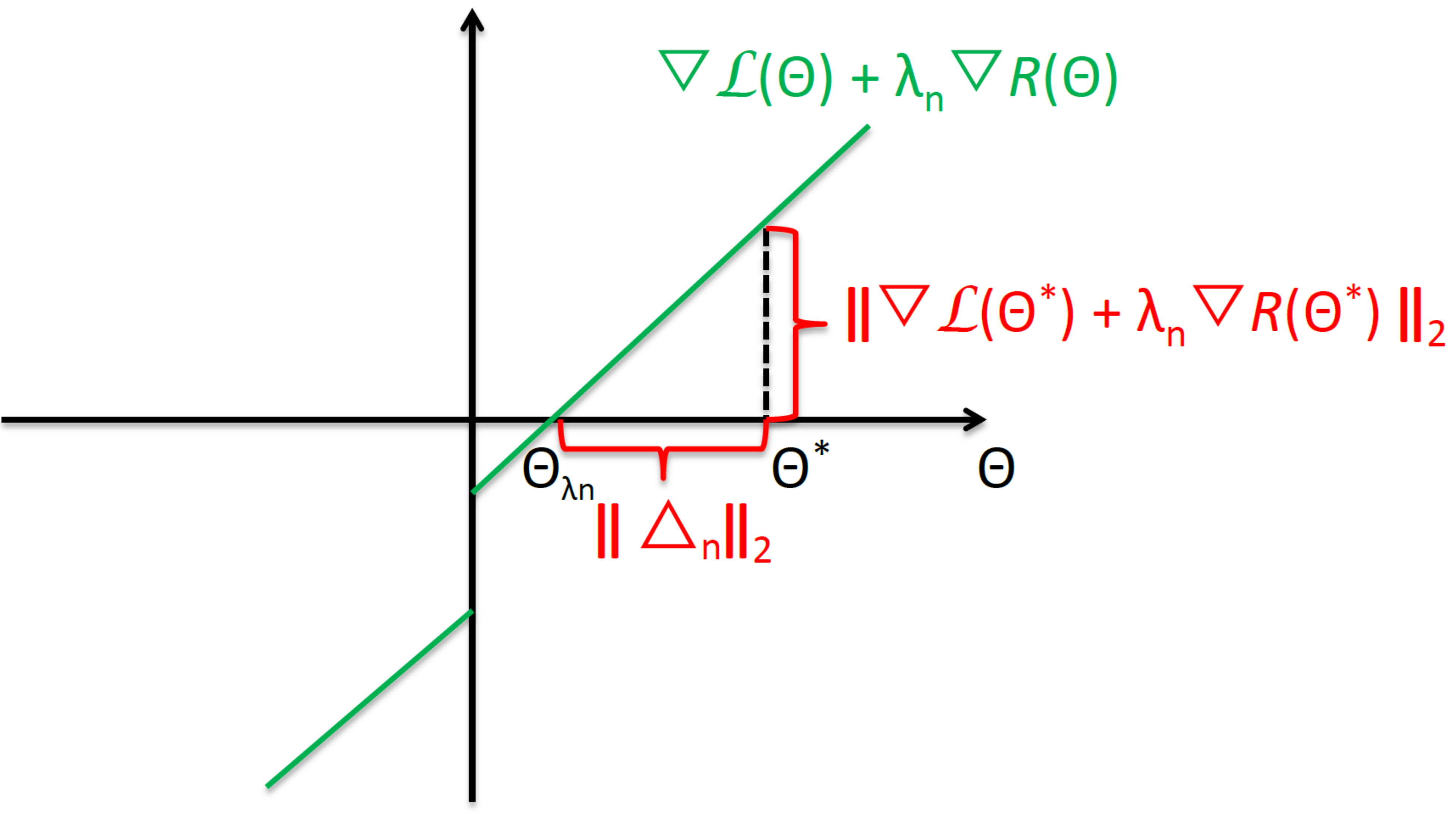}}
\caption{Schematic illustrating the error bound in Lemma \ref{lem:recbnderror}. Under restricted strong convexity (RSC) of the loss function in the error set $E_r$, the error $\| \hdn \|_2$ can be bounded in terms of the gradient of the overall objective evaluated at $\theta^*$.}
\label{fig:rsc}
\end{figure}

\subsection{A Special Case: Decomposable Norms}
In recent work, \cite{nrwy12} considered regularized regression with the special case of decomposable norms, defined in terms of a pair of subspaces ${\cal M} \subseteq \bar{{\cal M}}$ of $\R^p$. The model is assumed to be in the subspace ${\cal M}$, and the definition considers the so-called perturbation subspace $\bar{{\cal M}}^{\perp}$ which is the orthogonal complement of $\bar{{\cal M}}$. A norm $R(\cdot)$ is considered decomposable with respect to subspaces $({\cal M}, \bar{{\cal M}}^{\perp})$ if $R(\theta + \gamma) = R(\theta) + R(\gamma)$ for all $\theta \in {\cal M}$ and $\gamma \in \bar{{\cal M}}^{\perp}$.

We show that for decomposable norms, the error set $E_r$ in our analysis is included in the error cone defined in \cite{nrwy12}. In the current context, let $\beta = 2$,  $\theta^* \in \mathcal{M}$, then for any $\Delta = \Delta_{\bar{\mathcal{M}}^\perp} + \Delta_{\bar{\mathcal{M}}} \in E_r$, we have
\begin{eqnarray}
& R(\theta^* + \Delta) &\leq ~  R(\theta^*) + \frac{1}{2} R(\Delta) \\
\Rightarrow & \quad R(\theta^* + \Delta_{\bar{\mathcal{M}}^\perp} + \Delta_{\bar{\mathcal{M}}})  & \leq ~  R(\theta^*) + \frac{1}{2} R(\Delta_{\bar{\mathcal{M}}^\perp} + \Delta_{\bar{\mathcal{M}}}) \\
\Rightarrow & \quad R(\theta^* + \Delta_{\bar{\mathcal{M}}^\perp}) - R(\Delta_{\bar{\mathcal{M}}})  &\stackrel{(a)}{\leq} ~  R(\theta^*) + \frac{1}{2} R(\Delta_{\bar{\mathcal{M}}^\perp}) + \frac{1}{2}R(\Delta_{\bar{\mathcal{M}}}) \\
\Rightarrow & \quad R(\theta^*) + R(\Delta_{\bar{\mathcal{M}}^\perp}) - R(\Delta_{\bar{\mathcal{M}}})  &\stackrel{(b)}{\leq} ~  R(\theta^*) + \frac{1}{2} R(\Delta_{\bar{\mathcal{M}}^\perp}) + \frac{1}{2}R(\Delta_{\bar{\mathcal{M}}}) \\
\Rightarrow & \quad R(\Delta_{\bar{\mathcal{M}}^\perp})  &\leq ~  3 R(\Delta_{\bar{\mathcal{M}}}).
 \label{eq:decomErrSet}
\end{eqnarray}
where inequality (a) follows from the triangle inequality and (b) follows from decomposability of the norm. The last inequality is precisely the error cone in \cite{nrwy12} for $\theta^* \in \mathcal{M}$.
As a result, for any $\Delta \in E_r$, for decomposable norms we have
\begin{eqnarray}
R (\Delta) &=& R(  \Delta_{\bar{\mathcal{M}}^\perp} + \Delta_{\bar{\mathcal{M}}})
 \leq  R(  \Delta_{\bar{\mathcal{M}}^\perp}) + R(\Delta_{\bar{\mathcal{M}}})
\leq  4 R( \Delta_{\bar{\mathcal{M}}})
\end{eqnarray}
Hence, the norm compatibility constant can be bounded as
\beq
\psi(E_r) = \sup_{\Delta \in E_r} \frac{R(\Delta)}{\| \Delta \|_2}
 \leq 4 \sup_{\Delta \in E_r} \frac{R(\Delta_{\bar{\mathcal{M}}})}{\| \Delta \|_2}
 \leq 4 \sup_{\u \in \bar{\mathcal{M}} \setminus \{0\} } \frac{R(\u)}{\| \u \|_2} = 4 \Psi(\bar{\mathcal{M}}).
 \label{eq:relsubConstnormConst}
\eeq
where $\Psi(\bar{\mathcal{M}})$ is the subspace compatibility in $\bar{\mathcal{M}}$, as used in \cite{nrwy12}.


\section{Bounds on the Regularization Parameter}
\label{sec:lambda}

Recall that the parameter $\lambda_n$ needs to satisfy the inequality
\beq
\label{lambda_condition}
\lambda_n \geq \beta R^*(\nabla {\cal L}(\theta^*;Z^n))~.
\eeq
The right hand side of the inequality has two issues: the expression depends on $\theta^*$, the optimal parameter which is unknown, and the expression is a random variable, since it depends on $Z^n$. In this section, we characterize the expectation $E[R^*(\nabla {\cal L}(\theta^*;Z^n))]$ in terms of the Gaussian width of the unit norm ball $\Omega_R = \{ u : R(u) \leq 1\}$, and further discuss its upper bounds. For ease of exposition, we present results for the case of squared loss, i.e., ${\cal L}(\theta^*;Z^n) = \frac{1}{2n} \| y - X\theta^*\|^2$ with the linear model $y = X \theta + \omega$, where $\omega$ is noise vector with i.i.d. entries. Under this setting,
\beq
\nabla {\cal L}(\theta^*;Z^n) = \frac{1}{n} X^T (y - X\theta^*) = \frac{1}{n} X^T \omega~,
\eeq
which eliminates the dependency on the unknown $\theta^*$.
Before presenting the results, we introduce a few notations. We let $\Lambda_{\max} (\cdot)$ denote the largest eigenvalue of a square matrix. We also recall the definition of the sub-Gaussian norm for a sub-Gaussian variable $x$, $\vertiii{x}_{\psi_2} = \sup_{p \geq 1} \frac{1}{\sqrt{p}} ( E[|x|^p] )^{1/p}$~\cite{vers12}.

From this section onwards, the analysis will take into account the randomness of the design $X$ and the noise $\omega$. Here we give a brief description of our assumptions on $X$ and $\omega$ as follows, 

{\bf Isotropic Sub-Gaussian Designs:} the design matrix $X \in \R^{n \times p}$ has independent sub-Gaussian rows where each row satisfies $\vertiii{X_i}_{\psi_2} \leq \kappa$ and $E[X_i X_i^T] = \I_{p \times p}$. Thus, the measure $\mu$ from which the rows $X_i$ are sampled independently is an isotropic sub-Gaussian measure. \\
{\bf Anisotropic Sub-Gaussian Designs:} the design matrix $X \in \R^{n \times p}$ has independent rows, and each row $X_i$ is anisotropic sub-Gaussian with $E[X_i^T X_i] = \Sigma$. Further, we assume that corresponding isotropic random vector $\tilde{X}_i = X_i \Sigma^{-1/2}$ satisfies $\vertiii{\tilde{X}_i}_{\psi_2} \leq \kappa$. A simple special case of such an anisotropic sub-Gaussian design is when $X_i \sim N(0,\Sigma)$, where $\tilde{X}_i = X_i \Sigma^{-1/2} \sim N(0,\I)$ so that $\vertiii{\tilde{X}_i}_{\psi_2} = 1$. \\
{\bf Sub-Gaussian Noise:} the noise $\omega$ has i.i.d. centered unit-variance sub-Gaussian entries with $\vertiii{\omega_i}_{\psi_2} \leq K$. 

For convenience, we only use the shorthand in bold font to specify the assumptions. In the following theorem, we characterize the expectation of $R^*(\nabla {\cal L}(\theta^*;Z^n))$ in terms of Gaussian width of the unit norm ball $w(\Omega_R)$.
\begin{theo}
\label{theo:lambda_expect}
Let $\Omega_R = \{ u: R(u) \leq 1\}$, and $\cal L$ be the squared loss. For sub-Gaussian design $X$ and noise $\omega$, we have
\beq
E\left[R^*(\nabla {\cal L}(\theta^*;Z^n))\right] \leq \eta \xi \cdot \frac{\kappa w(\Omega_R)}{\sqrt{n}} ~,
\vspace*{-2mm}
\eeq
where the expectation is taken over both $X$ and $\omega$. The constant $\xi$ is given by
\begin{align*}
\xi = \left \{
             \begin{array}{lll}
              1 \ \ \ &\text{if $X$ is isotropic} \\
              \sqrt{\Lambda_{\max}(\Sigma)} \ \ \ &\text{if $X$ is anisotropic} ~. \\
             \end{array}  \right.
\end{align*}
\end{theo}
Bounding the expectation of $R^*(\nabla {\cal L}(\theta^*;Z^n))$ gives us a rough scale of the regularization parameter $\lambda_n$. In the next theorem, we present a high-probability upper bound for $R^*(\nabla {\cal L}(\theta^*;Z^n))$.
\begin{theo}
\label{theo:lambda_gauss_bound}
Let design $X$ and noise $\omega$ be sub-Gaussian, and $\cal L$ be squared loss. Define $\phi = \sup_{R(u) \leq 1} \|u\|_2$, then for any $\tau > 0$, with probability at least $1- c_1 \exp\left(- \min\left((\frac{\tau}{c_2 \xi \kappa \phi})^2, c_0 n\right)\right)$, we have
\beq
R^*\left(\nabla {\cal L}(\theta^*;Z^n)\right) \leq  \sqrt{\frac{2 K^2 + 1}{n}} \left(c \xi \kappa \cdot w(\Omega_R) + \tau\right) ~,
\label{eq:gausDualBound}
\eeq
where $c$, $c_0$, $c_1$ and $c_2$ are all absolute constants, and $\xi$ is the same as in Theorem \ref{theo:lambda_expect}.
\end{theo}
\begin{figure}
\centering
\includegraphics[width=0.6\textwidth]{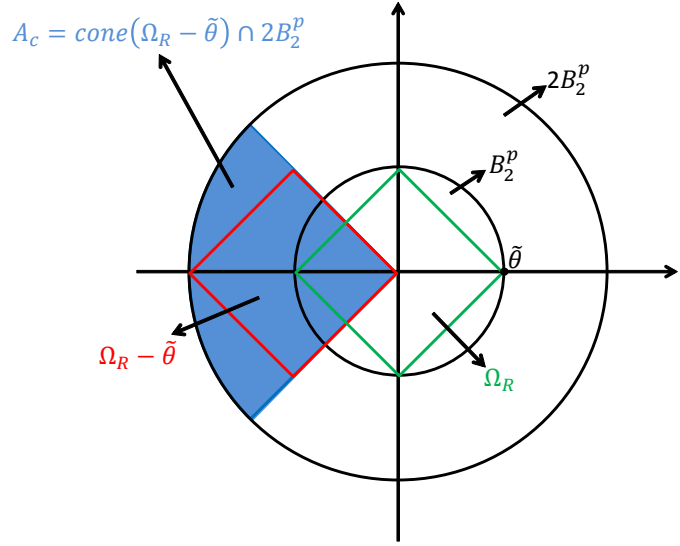}
\caption{Bounding the Gaussian width of a norm ball, e.g., corresponding to $L_1$ norm, by shifting the norm ball and using the width of the corresponding cone (Lemma \ref{lem:ballWidthBound}). The approach allows one to directly use existing results on bounding Gaussian widths of certain cones. In some cases, it may be easier to directly bound the Gaussian width of the norm ball, rather than using the shifting argument.}
\label{fig:gaussWidthL1Ball}
\end{figure}

{\bf Bounding the Gaussian width $w(\Omega_R)$:} In certain cases, one may be able to directly obtain a bound on the Gaussian width $w(\Omega_R)$. Here, we provide a mechanism for bounding the Gaussian width $w(\Omega_R)$ of the unit norm ball in terms of the Gaussian width of a suitable cone, obtained by shifting or translating the norm ball. In particular, the result involves taking any point on the boundary of the unit norm ball, considering that as the origin, and constructing a cone using the norm ball. Since such a construction can be done with any point on the boundary, the tightest bound is obtained by taking the infimum over all points on the boundary. The motivation behind getting an upper bound of the Gaussian width $w(\Omega_R)$ of the unit norm ball in terms of the Gaussian width of such a cone is because considerable advances have been made in recent years in upper bounding Gaussian widths of such cones~\cite{crpw12,almt14}.
\begin{lemm}
Let $\Omega_R = \{ u : R(u) \leq 1 \}$ be the unit norm ball and $\Theta_R = \{ u : R(u) = 1 \}$ be the boundary. For any $\tilde{\theta} \in \Theta_R$, $\rho(\tilde{\theta}) = \sup_{\theta:R(\theta) \leq 1} \|\theta - \tilde{\theta}\|_2$ is the diameter of $\Omega_R$ measured with respect to $\tilde{\theta}$. Let $G(\tilde{\theta}) = \cone(\Omega_R - \tilde{\theta}) \cap \rho(\tilde{\theta}) B_2^p$, i.e., the cone of $(\Omega_R - \tilde{\theta})$ intersecting the ball of radius $\rho(\tilde{\theta})$. Then
\beq
w(\Omega_R) \leq \inf_{\tilde{\theta} \in \Theta_R} ~w(G(\tilde{\theta}))~.
\eeq
\label{lem:ballWidthBound}
\end{lemm}
The analysis and results for $\lambda_n$ presented above can be extended to general convex losses arising in the context of GLMs for sub-Gussian designs and sub-Gaussian noise (see Section~\ref{sec:glm}).
.


\section{Least Squares Models: Restricted Eigenvalue Conditions}
\label{sec:re}

The error bound analysis in Theorem \ref{thm:recover} depends on the restricted strong convexity (RSC) assumption. In this section, we establish RSC conditions for sub-Gaussian design matrices when the loss function is the squared loss. For squared loss, i.e., $\cL(\theta; Z^n) = \frac{1}{n}\| \y - X \theta \|^2$, the RSC condition~\myref{eq:rsc} becomes equivalent to the Restricted Eigenvalue (RE) condition~\cite{birt09,nrwy12}, since
\beq
\begin{split}
\delta \cL(\Delta,\theta^*) & = \frac{1}{n} \| y - X (\theta^* + \Delta) \|^2 - \frac{1}{n} \| y - X \theta^* \|^2 + \frac{1}{n} \langle X^T (y - X \theta^*)  , \Delta \rangle \\
& = \frac{1}{n} \| X \Delta \|^2 = \frac{1}{n} \sum_{i=1}^n \langle X_i, \Delta \rangle^2~.
\end{split}
\eeq
so that the condition simplifies to
\beq
\delta \cL(\Delta,\theta^*) = \frac{1}{n} \sum_{i=1}^n \langle X_i, \Delta \rangle^2 \geq \kappa \| \Delta \|_2^2~,
\eeq
for all $\Delta \in E_r$.
We make two simplifications which lets us develop the RE results in terms of widths of spherical caps rather than over the error set $E_r$. Let $n_{E_r}$ be the sample complexity for the RE condition over the set $E_r$, so that for $n > n_{E_r}$ samples, with high probability
\beq
\underset{\Delta \in E_r}{\inf}~ \frac{1}{n} \| X \Delta \|_2^2 \geq \kappa_{E_r} \| \Delta \|_2^2~,
\eeq
for some $\kappa_{E_r} > 0$. Let $C_r = \cone(E_r)$ and let $n_{C_r}$ be the sample complexity for the RE condition over the cone $C_r$, so that for $n > n_{C_r}$ samples, with high probability
\beq
\underset{\Delta \in C_r}{\inf}~ \frac{1}{n} \| X \Delta \|_2^2 \geq \kappa_{C_r} \| \Delta \|_2^2~,
\eeq
for some $\kappa_{C_r} > 0$. Since $E_r \subseteq C_r$, we have $n_{E_r} \leq n_{C_r}$. Thus, it is sufficient to obtain (an upper bound to) the sample complexity $n_{C_r}$, since that will serve as an upper bound to $n_{E_r}$, the sample complexity over $E_r$. Further, since $C_r$ is a cone, the absolute magnitude $\| \Delta \|_2$ does not affect the sample complexity. As a result, it is sufficient to focus on a spherical cap $A = C_r \cap S^{p-1}$. In particular, if $n_A$ denotes the sample complexity for the RE condition over the spherical cap $A$, so that for $n > n_A$ samples, with high probability
\beq
\underset{u \in A}{\inf}~ \frac{1}{n} \| X u \|_2^2 \geq \bar{\kappa}_{A} \| u \|_2^2~,
\eeq
for some $\bar{\kappa}_A > 0$, then $n_A = n_{C_r} \geq n_{E_r}$. Noting that $\| u \|_2 = 1$ for $u \in C_r \cap S^{p-1}$, we consider sample complexity results for RE conditions the form
\beq
\underset{u \in A}{\inf}~ \frac{1}{n} \| X u \|^2_2 = \underset{u \in A}{\inf}~ \frac{1}{n} \sum_{i=1}^n \langle X_i, u \rangle^2 \geq \kappa_A(n,p)
\eeq
where $\kappa_A(n,p) > 0$ with high probability for $n > n_A$. In this section, we characterize sample complexity $n_A$ over {\em any} given spherical cap $A$, and establish RE conditions for isotropic and anisotropic sub-Gaussian design matrices $X$  in terms of the Gaussian width $w(A)$.

Analysis of RE conditions for certain types of design matrices for certain types of norms, especially the $L_1$ norm, have appeared in the literature~\cite{birt09,rawy10,ruzh13}. The RE/RIP conditions for independent isotropic Gaussian designs have been widely studied for the case of $L_1$ norm \cite{cand08,birt09}. The generalization to RE condition for correlated Gaussian designs for the special of $L_1$ norm was studied in~\cite{rawy10}. \cite{crpw12} consider the more general context of atomic norms, and RE condition analysis applies to any spherical cap $A$, with sample complexity results in terms of $w(A)$, the Gaussian width of $A$. However, the analysis relies on Gordon's inequality~\cite{gord85,gord88,ledo13}, which is applicable only for isotropic Gaussian design matrices. Progress has been made on establishing RE conditions for sub-Gaussian designs for error sets/caps corresponding to specific norms such as $L_1$ \cite{zhou09}. In recent work, RE conditions were developed for anisotropic sub-Gaussian designs for the $L_1$ norm~\cite{ruzh13}. Further, recent work have pointed out the differences between the RE and the RIP condition, which gives a two-sided bound on quadratic forms of random matrices~\cite{nrwy12}. In particular, while the RE condition is sufficient for structured estimation, the RIP results are stronger and may have higher sample complexity.

In the following, we establish the stronger RIP results for any spherical cap $A$ and any sub-Gaussian design matrix, handling the isotropic and anisotropic cases separately. The special case of Gaussian design matrices are automatically covered by the sub-Gaussian results, and results such as Gordon's inequality can be viewed as a special case. All results are in terms of $w(A)$, the Gaussian width of $A$, even for sub-Gaussian designs. In fact, all existing RE results do implicitly have the width term, but in a form specific to the chosen norm~\cite{rawy10,ruzh13}. The analysis on atomic norm in~\cite{crpw12} has the $w(A)$ term explicitly, but the analysis relies on Gordon's inequality~\cite{gord85,gord88,ledo13}, which is applicable only for isotropic Gaussian design matrices.

The proof technique we use is an application of generic chaining~\cite{tala05,tala14}. The specific form we utilize was originally developed in~\cite{klme05,mept07}. The main idea is to pose the RIP condition as a bound on the supremum of a suitable stochastic process, so that generic chaining can be invoked to obtain a bound. The key difference between our RIP analysis and much of the existing literature on RE conditions, which use specialized tools such as Gaussian comparison principles~\cite{rawy10,nrwy12} or analysis geared to a particular norm~\cite{ruzh13}, is the use of generic chaining which simplifies the analysis considerably. Further, the RIP results can be viewed as a generalization of the celebrated Johnson-Lindenstrauss (JL) lemma~\cite{dagu03}, and the interested reader can explore these connections in~\cite{klme05}.



{\bf Isotropic Sub-Gaussian Designs:} We first consider the case where the design matrix $X \in \R^{n \times p}$ has independent sub-Gaussian rows where each row satisfies $\vertiii{X_i}_{\psi_2} \leq \kappa$ and $E[X_i X_i^T] = \I_{p \times p}$. Thus, the measure $\mu$ from which the rows $X_i$ are sampled independently is an isotropic sub-Gaussian measure.
\begin{theo}
Let $X$ be a design matrix with independent isotropic sub-Gaussian rows, i.e., $\vertiii{X_i}_{\psi_2} \leq \kappa$ and $E[X_i X_i^T] = \I_{p \times p}$. Then, for absolute constants $\eta, c > 0$, with probability at least $(1-2 \exp(-\eta w^2(A)))$, we have
\beq
\sup_{u \in A} \left| \frac{1}{n} ||Xu||^2 - 1 \right| ~=~\sup_{u \in A} \left| \frac{1}{n} \sum_{i=1}^n \langle X_i, u \rangle^2 - 1 \right| ~\leq ~c\frac{w(A)}{\sqrt{n}}~,
\eeq
or, equivalently,
\beq
1 - c\frac{w(A)}{\sqrt{n}} ~~\leq ~~\inf_{u \in A}~ \frac{1}{n} ||Xu||^2 ~~\leq ~~\sup_{u \in A} ~\frac{1}{n} ||Xu||^2  ~~\leq ~~1 + c\frac{w(A)}{\sqrt{n}}~.
\eeq
\label{thm:isg-gc}
\end{theo}
As a result, for $n > c^2 w^2(A)$, the RE condition: $\inf_{u \in A}~\| X u \|^2 \geq 1 - c w(A)/\sqrt{n} > 0$ is satisfied with high probability for any sub-Gaussian design matrix. More generally, choosing $\epsilon = c w(A)/\sqrt{n}$, one can write the result in a traditional RIP form \cite{cand08}.

{\bf Anisotropic Sub-Gaussian Designs:}
We now consider the case where the design matrix $X \in \R^{n \times p}$ has independent rows, and each row $X_i$ is anisotropic sub-Gaussian with $E[X_i^T X_i] = \Sigma$. Further, we assume that corresponding isotropic random vector $\tilde{X}_i = X_i \Sigma^{-1/2}$ satisfies $\vertiii{\tilde{X}_i}_{\psi_2} \leq \kappa$. A simple special case of such an anisotropic sub-Gaussian design is when $X_i \sim N(0,\Sigma)$, where $\tilde{X}_i = X_i \Sigma^{-1/2} \sim N(0,\I)$ so that $\vertiii{\tilde{X}_i}_{\psi_2} = 1$. The result below characterizes RIP-style property of any such anisotropic sub-Gaussian designs.
\begin{theo}
Let $X$ be a design matrix with independent anisotropic sub-Gaussian rows, i.e., $E[X_i^T X_i] = \Sigma$ and $\vertiii{X_i \Sigma^{-1/2}}_{\psi_2} \leq \kappa$. Then, for absolute constants $\eta, c > 0$, with probability at least $(1-2 \exp(-\eta w^2(A)))$, we have
\beq
\sup_{u \in A} \left| \frac{1}{n} \frac{1}{u^T \Sigma u} ||Xu||^2 - 1 \right| ~=~\sup_{u \in A} \left| \frac{1}{n} \frac{1}{u^T \Sigma u} \sum_{i=1}^n \langle X_i, u \rangle^2 - 1 \right| ~\leq ~c\frac{w(A)}{\sqrt{n}}~.
\eeq
Further,
\beq
\lambda_{\min}(\Sigma|A)\left( 1 - c\frac{w(A)}{\sqrt{n}} \right) ~~\leq ~~\inf_{u \in A}~ \frac{1}{n} ||Xu||^2 ~~\leq ~~\sup_{u \in A} ~\frac{1}{n} ||Xu||^2  ~~\leq ~~\lambda_{\max}(\Sigma|A) \left(1 + c\frac{w(A)}{\sqrt{n}} \right)~,
\eeq
where
\beq
\lambda_{\min}(\Sigma|A) = \inf_{u \in A}~u^T \Sigma u~, \qquad \text{and} \qquad \lambda_{\max}(\Sigma|A) = \sup_{u \in A}~u^T \Sigma u~
\eeq
are the restricted minimum and maximum eigenvalues of $\Sigma$ restricted to $A \subseteq S^{p-1}$.
\label{thm:iasg-gc}
\end{theo}
Thus, for the anisotropic case, the RIP is with respect to the restricted minimum and maximum eigenvalues corresponding to $A \subseteq S^{p-1}$. For the special case when $A = S^{p-1}$, we have $\lambda_{\min}(\Sigma|A) = \lambda_{\min}(\Sigma)$, the minimum eigenvalue, and $\lambda_{\max}(\Sigma|A) = \lambda_{\max}(\Sigma)$, the maximum eigenvalue of $\Sigma$. Further, when $\Sigma = \I$, we get back the result in Theorem~\ref{thm:isg-gc}. Finally, it is instructive to compare the above result to existing characterizations of the RE condition for anisotropic Gaussian \cite{rawy10} and anisotropic sub-Gaussian ~\cite{ruzh13} designs, focused on the $L_1$ norm. The result in Theorem~\ref{thm:iasg-gc} is a more general RIP result, applies to any spherical cap $A$, and  is in terms of $w(A)$, the Gaussian width of the spherical cap $A$. 

\section{Examples and Applications}
\label{sec:examples}
In this section, we give examples of the analysis from previous sections for three norms: $L_1$ norm, group sparse norm, and $L_2$ norm. The summary of the results is given in Table \ref{tab:recBndGauss}. Other examples can be constructed for norms and error sets with known bounds on the Gaussian widths and norm compatibility constants~\cite{crpw12,nrwy12}, and more general ways to bound the Gaussian widths and norm compatibility constants have been developed in \cite{mapr14,chba15}.


\textbf{$\mathbf{L_1}$ Norm:}
Assume that the statistical parameter $\theta^*$ is $s$-sparse, and note that $\| \theta^*\|_1 \leq \sqrt{s} \| \theta^*\|_2$. 
Since $L_1$ norm is a decomposable norm following the result in \eqref{eq:decomErrSet},
we have $\Psi(E_r) \leq 4 \Psi(\bar{\mathcal{M}}) = 4\sqrt{s}$.

Applying Lemma \ref{lem:ballWidthBound}, let $\bar{\theta}$ be a 1-sparse vector, and $\rho(\bar{\theta}) = 2$, then $w(\Omega_R)$ can be bounded by
\beq
w(\Omega_R) \leq \inf_{\tilde{\theta} \in \Theta_R} w(G(\tilde{\theta})) = w(G(\bar{\theta})) \overset{(a)}{=} O \left ( \sqrt{\log p} \right ),
\label{eq:bndOmegaRLasso}
\eeq
where (a) is obtained from the fact that Gaussian width of $G(\tilde{\theta})$ with $\tilde{\theta}$ be a s-sparse vector is $\sqrt{2s \log(\frac{p}{s})+ \frac{5}{4}s}$ \cite{crpw12}. See Figure \ref{fig:gaussWidthL1Ball} for more details.
From Theorem \ref{theo:lambda_gauss_bound} and \eqref{eq:bndOmegaRLasso}, the bound on $\lambda_n$ is
\beq
\lambda_n \leq c \frac{w(\Omega_R)}{\sqrt{n}} = O \left ( \sqrt{\frac{\log p}{n}} \right ).
\eeq
Hence, the recovery error is bounded by
\beq
\| \hat{\Delta}_n \|_2 \leq c_3 \frac{\Psi(E_r) \lambda_n }{\kappa} = O \left ( \sqrt{\frac{s \log p}{n}} \right ),
\eeq
which is similar to the results obtained in well known results \cite{crpw12, nrwy12}.

\textbf{Group Sparse Norm:}
Suppose that the index set $\{1, 2, \cdots, p\}$ can be partitioned into a set of $T$ disjoint
groups, say $\mathcal{G} = \{\mathcal{G}_1, \mathcal{G}_2, \cdots , \mathcal{G}_T \}$.  Define $(1, \nu)$-group norm for a  given vector $\boldsymbol{\nu} = (\nu_1, \cdots, \nu_T ) \in [1,\infty]^T$ as
\beq
\| \alpha \|_{\mathcal{G}, \boldsymbol{\nu}} = \sum_{t=1}^T \| \alpha_{\mathcal{G}_t} \|_{\nu_t}
\eeq
As shown in \cite{nrwy12} Group norm is a decomposable norm. For a given subset $S_\mathcal{G} \subset \{1, \ldots, T \}$ with cardinality $|S_\mathcal{G}|$, define the subspace $A(S_\mathcal{G}) = \{ \alpha \in \mathbb{R}^p \left| \alpha_{\mathcal{G}_t} = 0, ~~ \forall t\notin S_\mathcal{G} \right.\} $. Let $\nu_t \geq 2$, then we have
\beq
\| \Delta \|_{\mathcal{G}, \boldsymbol{\nu}} = \sum_{t \in S_\mathcal{G}} \| \Delta_{\mathcal{G}_t} \|_{\nu_t} \leq \sum_{t \in S_\mathcal{G}} \| \Delta_{\mathcal{G}_t} \|_2 \leq \sqrt{s_\mathcal{G}} \| \Delta \|_2.
\label{eq:subspaceConstGrpLas}
\eeq
Hence, from \eqref{eq:relsubConstnormConst} and \eqref{eq:subspaceConstGrpLas} we have
\beq
\Psi(E_r) \leq 4 \sqrt{s_\mathcal{G}}.
\eeq

Applying Lemma \ref{lem:ballWidthBound}, define $\bar{\theta}$ with 1-active group, and $\rho(\bar{\theta}) = 2$, then $w(\Omega_R)$ can be bounded by
\beq
w(\Omega_R) \leq \inf_{\tilde{\theta} \in \Theta_R} w(G(\tilde{\theta})) = w(G(\bar{\theta})) \overset{(a)}{=} O \left ( \sqrt{m + \log T} \right ),
\label{eq:bndOmegaRGrpLasso}
\eeq
where $m = \underset{t}{\max} |\mathcal{G}_t|$ and (a) is obtained from the fact that Gaussian width of $G(\tilde{\theta})$ where $\tilde{\theta}$ has $k$ active group is $\sqrt{2k (m + \log(T - k)) + k} $ \cite{crpw12}.
From Theorem \ref{theo:lambda_gauss_bound} and \eqref{eq:bndOmegaRGrpLasso}, the bound on $\lambda_n$ is
\beq
\lambda_n \leq c \frac{w(\Omega_R)}{\sqrt{n}} = O \left ( \sqrt{\frac{m + \log T}{n}} \right ).
\eeq
Hence, the recovery error is bounded by
\beq
\| \hat{\Delta}_n \|_2 \leq c_3 \frac{\Psi(E_r) \lambda_n }{\kappa} = O \left ( \sqrt{\frac{s_\mathcal{G} (m+\log T)}{n}} \right ),
\eeq
which is similar to the results obtained in previous works \cite{crpw12, nrwy12}.



\textbf{$\mathbf{L_2}$ Norm:}
With $L_2$ norm as the regularizer, the norm constant is obtained as
\beq
\Psi(E_r) = \sup_{\Delta \in E_r} \frac{\|  \Delta \|_2}{\| \Delta \|_2} = 1.
\eeq
Applying Lemma \ref{lem:ballWidthBound}, set $\rho(\tilde{\theta}) = 1$, then $w(\Omega_R)$ can be bounded by
\beq
w(\Omega_R) \leq \inf_{\tilde{\theta} \in \Theta_R} w(G(\tilde{\theta})) =  O \left ( \sqrt{p} \right ).
\label{eq:bndOmegaRL2}
\eeq
From Theorem \ref{theo:lambda_gauss_bound} and \eqref{eq:bndOmegaRL2}, the bound on $\lambda_n$ is
\beq
\lambda_n \leq c \frac{w(\Omega_R)}{\sqrt{n}} = O \left ( \sqrt{\frac{p}{n}} \right ).
\eeq
Hence, the recovery error is bounded by
\beq
\| \hat{\Delta}_n \|_2 \leq c_3 \frac{\Psi(E_r) \lambda_n }{\kappa} = O \left ( \sqrt{\frac{p}{n}} \right ).
\eeq

\begin{table}
\begin{center}
{\small
\begin{tabular}{c|c|c|c|c}
$R(u)$ & $\lambda_n:=c_1 \frac{w(\Omega_R)}{\sqrt{n}}$ & $\kappa :=  \left [ \max \left \{ \left ( 1 - \sqrt{c_2} \frac{w(A)}{\sqrt{n}} \right ), 0 \right \} \right ]^2$  & $\Psi(E_r)$ & $\| \hat{\Delta}_n \|_2 := c_3 \frac{\Psi(E_r) \lambda_n }{\kappa}$ \\
\hline
\hline
&&&& \\
$L_1$ & $O \left ( \sqrt{\frac{\log p}{n}} \right )$ & $\Theta( 1 )$ if $n > c_2w^2(A) = O(s \log p)$ & $\sqrt{s}$ & $O \left ( \sqrt{\frac{s \log p}{n}} \right )$ \\
&&&& \\
Group sparse & $O \left ( \sqrt{\frac{m + \log T}{n}} \right )$ & $\Theta(1)$ if $n > c_2w^2(A) = O(s_{\mathcal{G}}(m + \log T))$ & $\sqrt{s_\mathcal{G}}$ & $O \left ( \sqrt{\frac{s_\mathcal{G} (m+\log T)}{n}} \right )$\\
&&&& \\
$L_2$  & $O \left ( \sqrt{\frac{p}{n}} \right )$ & $\Theta(1)$ if $n > c_2w^2(A)=O(p)$ & $1$ & $O \left ( \sqrt{\frac{p}{n}} \right )$ \\
\end{tabular}}
\end{center}
\caption{A summary of values for the regularization parameter $\lambda_n$, the RE condition constant $\kappa$, the norm constant $\Psi(E_r)$ and recovery bounds $\| \hat{\Delta}_n \|_2$ for $\ell_1$, $\ell_2$ and group norms in case of Gaussian Design matrix with Gaussian noise. All results are given up to constants with more emphasis on the scale of the results.}
\label{tab:recBndGauss}
\end{table}

\section{Generalized Linear Models: Restricted Strong Convexity}
\label{sec:glm}

In this section, we extend our results to estimation with norm regularization in the context of generalized linear models (GLMs)~\cite{barn78,brow86}. Assume that the conditional distribution of the response $y_i$ conditioned on the covariates $X_i$ is an exponential family distribution:
\beq
p(y_i|X_i;\theta^*) = p(y_i|\langle X_i,\theta^*\rangle) = \exp \{ y_i \langle X_i, \theta^* \rangle - \varphi(\langle X_i,\theta^* \rangle) \}~.
\eeq
where
\beq
\varphi(\langle X_i,\theta^* \rangle) = \log \left( \int_{y_i} \exp \{ y_i \langle X_i, \theta^* \rangle\}~ dy_i \right)~
\eeq
is the log-partition function~\cite{brow86,barn78,wajo08}.\footnote{Note that for GLMs over discrete responses $y_i$, the integration needs to be suitably changed to summation~\cite{barn78,brow86}.} In GLMs, the conditional distribution of the response $y_i$ is characterized by an exponential family distribution $p(y_i|\eta_i)$ with natural parameter $\eta_i = \langle X_i, \theta^* \rangle$ determined by the covariates $X_i$ and the parameter $\theta^*$. It is easy to verify that the gradient of the log-partition function w.r.t.~the natural parameter $\eta_i =  \langle X_i, \theta^* \rangle$ gives the expectation of the response~\cite{barn78,brow86}, i.e.,
\beq
\nabla_{\eta_i} \varphi(\eta_i) = \nabla_{\langle X_i, \theta^* \rangle} \varphi(\langle X_i, \theta^* \rangle) = E[y_i|\langle X_i, \theta^* \rangle]~.
\eeq
For estimating $\theta^*$, the loss function corresponding to GLMs typically consider the
negative log likelihood of the conditional distribution:
\beq
\cL(\theta;Z^n) =  - \frac{1}{n} \log p(y_i|X_i;\theta^*) = \frac{1}{n} \sum_{i=1}^n ( \varphi(\langle X_i, \theta \rangle) -  y_i \langle X_i, \theta \rangle )~.
\eeq
In the current context, we assume $\theta*$ to be sparse/structured, and the structure can be suitably captured by a norm $R(\cdot)$. Then, the estimation of $\theta^*$ with norm regularization takes the form:
\beq
\hat{\theta}_{\lambda_n} = \underset{\theta \in \R^p}{\argmin} ~\cL(\theta;Z^n) + \lambda_n R(\theta)
= \underset{\theta \in \R^p}{\argmin} ~ \frac{1}{n} \sum_{i=1}^n ( \varphi(\langle X_i, \theta \rangle) -  y_i \langle X_i, \theta \rangle ) + \lambda_n R(\theta)~.
\eeq
Noise in the context of GLMs is simply the deviation of a specific response $y_i$ from the conditional mean, i.e., $\omega_i = E[y_i|X_i] - y_i$. Popular examples of GLMs come from suitable choices of the conditional distribution, e.g., when $p(y_i|\langle X_i, \theta \rangle)$ is Gaussian so that $\varphi(\langle X_i,\theta \rangle ) = \frac{\langle X_i,\theta \rangle^2}{2}$, Bernoulli so that $\varphi(\langle X_i,\theta \rangle ) = \log(1+\exp(\langle X_i,\theta \rangle))$, and Poisson where $\varphi(\langle X_i,\theta \rangle) = \exp(\langle X_i,\theta \rangle)$, respectively yielding  least squares regression, logistic regression, and Poisson regression loss functions. Next, we provide the key results needed to characterize the regularization parameter $\lambda_n$  and restricted strong convexity (RSC) in the context of GLMs. The non-asymptotic bound on the estimation error then follows from the general result in \myref{eq:recover}.

{\bf Bounds on the Regularization Parameter:}
Following the general analysis from Section~\ref{sec:lambda}, the regularization parameter needs to satisfy the condition: $\lambda_n \geq \beta R^*(\nabla_{\theta} \cL (\theta^*;Z^n))$ for any fixed $\beta > 1$. For GLMs,
\beq
\nabla_{\theta} \cL(\theta^{*};Z^{n}) = -\frac{1}{n} \sum_{i=1}^{n} y_{i}X_{i} + \frac{1}{n} \sum_{i=1}^{n} X_{i} \nabla_{\langle X_i,\theta^* \rangle} \varphi(\langle X_i,\theta^* \rangle) = \frac{1}{n} \sum_{i=1}^n X_i(E[y|X_i] - y_i) = \frac{1}{n}X^{T} \omega ~,
\eeq
where we have used the fact, $\nabla_{\langle X_i,\theta^* \rangle} \varphi(\langle X_i,\theta^* \rangle) = E[y_i|\langle X_i,\theta^* \rangle]$ and $\omega = E[y_i|\langle X_i,\theta^* \rangle] - y_i$. Thus, the form of $\nabla_{\theta} \cL(\theta^{*};Z^{n})$ is the same as that in Section~\ref{sec:lambda}. Assuming the design matrix $X$ and noise $\omega$ are sub-Gaussian, a characterization of $\lambda_n$ follows from Theorems~\ref{theo:lambda_expect} and \ref{theo:lambda_gauss_bound} in Section~\ref{sec:lambda}. In particular,
$E[\nabla_{\theta} \cL(\theta^{*};Z^{n})] = O( \frac{w(\Omega_R)}{\sqrt{n}})$, with corresponding high probability concentration results, and it suffices to have $\lambda_n$ to be of this order.

{\bf Restricted Strong Convexity:}
By definition, the restricted strong convexity considers
\begin{align*}
\delta \cL(u,\theta^*) = \cL(\theta^*+ u) - \cL(\theta^*) - \langle \nabla \cL(\theta^*),u \rangle
    = \frac{1}{n} \sum_{i=1}^{n} \nabla^2 \varphi (\langle \theta^{*}, X_{i} \rangle + \gamma_{i} \langle u, X_i \rangle) \langle u,X_i \rangle ^{2}~,
\end{align*}
where $\gamma_{i} \in [0, 1]$, and where the last equality follows from a direct application of the mean value theorem~\cite{rudi76}. Since the log-partition function $\varphi$ is of Legendre type~\cite{bmdg05,barn78,brow86}, the second derivative $\nabla^2 \varphi(\cdot)$ is always positive. Since the RSC condition relies on a non-trivial lower bound for the above quantity, the analysis will consider suitable compact sets where $\nabla^2 \varphi(\cdot)$ is bounded away from zero by a constant. In particular, for a suitable constant $T$, we consider the sets $\{ X_i | | \langle X_{i}, \theta^{*} \rangle| < T\}$ and $\{ X_i | | \langle X_{i},u \rangle | < T \}$. For $X_i$ lying in these sets, the argument $a = \langle X_{i}, \theta^{*} \rangle + \gamma_{i} \langle u, X_i \rangle$ of the second derivative satisfies $|a| \leq 2T$, which is the compact set of interest. Within the compact set,
$\ell = \ell_{\varphi}(T) = \min_{|a| \leq 2T} \nabla^2 \varphi(a)$ is bounded away from zero. Outside the compact set, we will only assume $\nabla^2\varphi(\cdot) > 0$. Based on the above construction, we have
\beq
\delta \cL (u,\theta^{*}) \geq \frac{\ell}{n} \sum_{i=1}^{n} \langle X_{i},u \rangle^{2} ~ \mathbb{I}[| \langle X_{i}, \theta^{*} \rangle| < T] ~\mathbb{I}[| \langle X_{i},u \rangle | < T]~.
\eeq
The quadratic form based lower bound allows us to establish RSC conditions for GLMs with isotropic subGaussian design matrices by building on results from Section~\ref{sec:re} for RE conditions for squared loss. As a result, the sample complexity of the RSC condition is also expressed in terms of the Gaussian width of the spherical cap $A$ derived from the error set. The analysis can be suitably generalized to anisotropic design matrices using techniques discussed in Section~\ref{sec:re}.

As before, we consider $u \in A \subseteq S^{p-1}$ so that $\|u\|_2 = 1$. Assuming $X$ has isotropic sub-Gaussian rows with $\vertiii{X_i}_{\varphi_2} \leq \kappa$, $\langle X_{i}, \theta^{*} \rangle$ and $\langle X_{i}, u \rangle$ are sub-Gaussian random variables with sub-Gaussian norm at most $C\kappa$~\cite{vers14}. Denote by $\varepsilon_1$ and $\varepsilon_2$ the probability that $\langle X_i,u \rangle$ and $\langle X_i,\theta^* \rangle$ exceeds some constant $T$, i.e., $\varepsilon_1(T;u) = P\{ |\langle X_i, u \rangle| > T \} \leq e \cdot \exp( -c_{2}T^2/C^2 \kappa^2) = \bar{\varepsilon}_1$, and $\varepsilon_2(T;\theta^*) = P\{ | \langle X_i, \theta^* \rangle | > T \} \leq e \cdot \exp(- c_2 T^2/C^2 \kappa^2) = \bar{\varepsilon}_2$, where $\bar{\varepsilon}_1 = \bar{\varepsilon}_1(T,\kappa)$ and $\bar{\varepsilon}_2 = \bar{\varepsilon}_2(T;\kappa)$ are uniform upper bounds on the individual tail probabilities. The result we present below is in terms of the above defined constants $\ell = \ell_{\varphi}(T)$, $\bar{\varepsilon}_1 = \bar{\varepsilon}_1(T,\kappa)$ and $\bar{\varepsilon}_2 =  \bar{\varepsilon}_2(T,\kappa)$ for any suitably chosen $T$.
\begin{theo}
Let $X \in \R^{n \times p}$ be a design matrix with independent isotropic sub-Gaussian rows such that $\vertiii{X_i}_{\varphi_2} \leq \kappa$. Then, for any set $A \subseteq S^{p-1}$ for suitable constants $\eta,c > 0$, with probability at least $1- 2 \exp \left(-\eta w^2(A) \right)$, we have
\beq
\inf_{u \in A} \partial \cL (u,\theta^{*}) \geq \ell \underline{\rho}^2 \left (1 - c\kappa_1^2 \frac{w(A)}{\sqrt{n}} \right ) ~,
\eeq
where $\underline{\rho}^2 = \inf_{u \in A} \rho_u^2$, with $\rho_u^2 = E[\langle X_{i},u \rangle^{2} \mathbb{I}[| \langle X_{i}, \theta^{*} \rangle| < T] \mathbb{I}[| \langle X_{i},u \rangle | < T ]]$, and $\kappa_1 = \frac{\kappa}{1-\bar{\varepsilon}_1 - \bar{\varepsilon}_2}$.
\label{thm:rscsubg}
\end{theo}
The form of the result is closely related to the corresponding result for the RE condition $\inf_{u \in A} \|X u\|_2$ considered in Section~\ref{sec:re}. Note that RSC analysis for GLMs was considered in~\cite{nrwy12} for specific norms, especially $L_1$, whereas our analysis applies to any set $A \subseteq S^{p-1}$, hence to any norm, and the result is in terms of the Gaussian width $w(A)$ of $A$.
Further, following arguments in Section~\ref{sec:re}, the RE analysis for GLMs can be extended to anisotropic subGaussian design matrices.

\section{Conclusions}
\label{sec:conc}
The paper presents a general set of results and tools for characterizing non-asymptotic estimation error in norm regularized regression problems. The analysis holds for any norm, and subsumes much of existing literature focused on structured sparsity and related themes. The work can be viewed as a direct generalization of results in~\cite{nrwy12}, which presented related results for decomposable norms. Our analysis illustrates the important role Gaussian widths, as a measure of size of suitable sets, play in such results. Further, the error sets for regularized and constrained versions of such problems are shown to be closely related~\cite{birt09}.

While the paper presents a unified geometric treatment of non-asymptotic structured estimation with regularized estimators, several technical questions need further investigation. The focus of the analysis has been on thin-tailed distributions, and the RE/RSC type analysis presented really gives two sided bounds, i.e., RIP, showing that thin-tailed distributions do satisfy the RIP condition. For heavy tailed measurements, the lower and upper tails of quadratic forms behave differently~\cite{oliv13,mend14}, and it may be possible to establish geometric estimation error analysis for general norms, some special cases of which have been investigated in recent years~\cite{kome13,leme14,mend14}. Further, the sample complexity of the phase transitions in the RE/RSC conditions for anisotropic designs depend on the largest eigenvalue (operator norm) of the covariance matrix, making the estimator sample inefficient for highly correlated designs. Since real-world several problems, including spatial and temporal problems, do have correlated observations, it will be important to investigate estimators which perform well in such settings~\cite{fino14}. Finally, the focus of the work is on parametric estimation, and it will be interesting to explore generalizations of the analysis to non-parametric settings. 


\section*{Appendix}

\appendix

\section{Background and Preliminaries}
\label{app:back}
We start with a review of some definitions  and well-known results which will be used for our proofs.

\subsection{Gaussian Width}
In several of our proofs, we use the concept of Gaussian width~\cite{gord88,crpw12}, which is defined as follows.

\begin{defn}[Gaussian width]
For any set $A \in \mathbb{R}^p$, the \emph{Gaussian width} of the set $A$ is defined as:
\beq
w(A) = E_g\left[\sup_{u \in A} \langle g, u \rangle\right]~,
\eeq
where the expectation is over $g \sim N(0,\I_{p \times p})$, a vector of independent zero-mean unit-variance Gaussian random variable.
\label{def:gw}
\end{defn}
The Gaussian width $w(A)$ provides a geometric characterization of the size of the set $A$. We consider three perspectives of the Gaussian width, and provide some properties which are used in our analysis.
First, consider the Gaussian process $\{ Z_u \}$  where the constituent Gaussian random variables $Z_u = \langle t, g \rangle$ are indexed by $u \in A$, and $g \sim N(0,\I_{p \times p})$. Then the Gaussian width $w(A)$ can be viewed as the expectation of the supremum of the Gaussian process $\{ Z_t \}$. Bounds on the expectations of Gaussian and other empirical processes have been widely studied in the literature, and we will make use of generic chaining for some of our analysis \cite{tala05,tala14,bolm13,ledo13}.
Second, $\langle u, g \rangle$ can be viewed as a Gaussian random projection of each $u \in A$ to one dimension, and the Gaussian width simply measures the expectation of largest value of such projections.
Third, if $A$ is the unit ball of any norm $R(\cdot)$, i.e., $A = \{ x \in \mathbb{R}^p \ | \ R(x) \leq 1\}$, then $w(A) = E_g[R^*(g)]$ by definition of the dual norm. Thus, the Gaussian width is the expected value of the dual norm of a standard Gaussian random vector. For instance, if $A$ is unit ball of $L_1$ norm, $w(A) = E [\|g\|_{\infty}]$.


Below we list some simple and useful properties of the Gaussian width of $A \subseteq \mathbb{R}^p$: \\
\\
{\bf Property 1:} $w(A) \leq w(B)$ for $A \subseteq B$. \\
\\
{\bf Property 2:} $w(A) = w(\text{conv}(A))$, where $\text{conv}(\cdot)$ denotes the convex hull of $A$. \\
\\
{\bf Property 3:} $w(cA) = c w(A)$ for any positive scalar $c$, in which $cA = \{cx \ | \ x \in A\}$. \\
\\
{\bf Property 4:} $w(\Gamma A) = w(A)$ for any orthogonal matrix $\Gamma \in \mathbb{R}^{p \times p}$. \\
\\
{\bf Property 5:} $w(A+b) = w(A)$ for any $A \subseteq \mathbb{R}^p$ and fixed $b \in \mathbb{R}^p$.

The last two properties illustrate the Gaussian width is rotation and translation invariant.

\subsection{Sub-Gaussian and Sub-exponential Random Variables (Vectors)}
In the proof, we will also frequently use the properties of sub-Gaussian and sub-exponential random variables (vectors). In particular, we are interested in their definitions using moments.
\begin{defn}
{\bf Sub-Gaussian (sub-exponential) random variable:} We say that a random variable $x$ is sub-Gaussian (sub-exponential) if the moments satisfies
\beq
[E|x|^p]^{\frac{1}{p}} \leq K_2 \sqrt{p} \ \ \ ( [E|x|^p]^{\frac{1}{p}} \leq K_1 p )
\eeq
for any $p \geq 1$ with a constant $K_2$ ($K_1$). The minimum value of $K_2$ ($K_1$) is called sub-Gaussian (sub-exponential) norm of $x$, denoted by $\vertiii{x}_{\psi_2}$ ($\vertiii{x}_{\psi_1}$).
\end{defn}

\begin{defn}
{\bf Sub-Gaussian (sub-exponential) random vector:} We say that a random vector $X$ in $\R^n$ is sub-Gaussian (sub-exponential) if the one-dimensional marginals $\langle X,x \rangle$ are sub-Gaussian (sub-exponential) random variables for all $x \in \R^n$. The sub-Gaussian (sub-exponential) norm of $X$ is defined as
\beq
\vertiii{X}_{\psi_2} = \sup\limits_{x \in S^{n-1}} \|\langle X, x \rangle \|_{\psi_2} \ \ \ (\vertiii{X}_{\psi_1} = \sup\limits_{x \in S^{n-1}} \|\langle X, x \rangle \|_{\psi_1})
\eeq
\label{def:subgvec}
\end{defn}

The following definitions and lemmas are from~\cite{vers12}.

\begin{lemm}
Consider a finite number of independent centered sub-Gaussian random variables $X_i$. Then $\sum_{i}X_i$ is also a centered sub-Gaussian random variable. Moreover,
\beq
\vertiii{\sum_{i}X_i}^{2}_{\psi_2} \leq C \sum_{i}\vertiii{X_i}_{\psi_2}^2
\eeq
\label{lem:rotinvsubg}
\end{lemm}

%

\begin{lemm}
Let $X_1, \ldots, X_n$ be independent centered sub-Gaussian random variables. Then $X = (X_1, \ldots, X_n)$ is a centered sub-Gaussian random vector in $\R^n$, and
\beq
\vertiii{X}_{\psi_2} \leq C \max\limits_{i \leq n} \vertiii{X_i}_{\psi_2}
\eeq
where $C$ is an absolute constant.
\label{lem:prosubg}
\end{lemm}
%
%

\begin{lemm}
Consider a sub-Gaussian random vector $X$ with sub-Gaussian norm $K = \max_{i} \vertiii{ X_i }_{\psi_2}$, then, $Z = \langle X, a\rangle$ is a sub-Gaussian random variable with sub-Gaussian norm $\vertiii{ Z }_{\psi_2} \leq C K \| a \|_2$.
\label{lem:rotsubGau}
\end{lemm}

\begin{lemm}
A random variable $X$ is sub-Gaussian if and only if $X^2$ is sub-exponential. Moreover,
\beq
\vertiii{X}^2_{\psi_{2}} \leq \vertiii{X^2}_{\psi_{1}} \leq 2\vertiii{X}_{\psi_2}^2
\eeq
\end{lemm}

\begin{lemm}
If $X$ is sub-Gaussian (or sub-exponential), then so is $X - EX$. Moreover, the following holds,
\beq
\vertiii{X - EX}_{\psi_2} \leq 2 \vertiii{X}_{\psi_2}, ~~~ \vertiii{X - EX}_{\psi_1} \leq 2 \vertiii{X}_{\psi_1}
\eeq
\end{lemm}

\section{Restricted Error Set and Recovery Guarantees}

Section \ref{sec:error} is about the restricted error set. Lemma \ref{lem:reserrset} characterizes the restricted error set. Theorem \ref{thm:relconsreg} establishes the relation between the constrained and restricted error sets. In particular, we prove that the Gaussian width of the regularized and constrained error sets (cone) are of the same order. Starting with the assumption that the RSC condition is satisfied Lemma \ref{lem:recbnderror} and Theorem \ref{thm:recover} derive results on the upper bound on the $L_2$ norm of the error.

We collect the proofs of the different results in this section.

\subsection{The Restricted Error Set}
Lemma \ref{lem:reserrset} in Section \ref{sec:error} characterizes the set to which the error vector belongs. We give the proof of Lemma \ref{lem:reserrset} below:

\textbf{Lemma \ref{lem:reserrset}}
\textit{
For any $\beta > 1$, assuming
\beq
\lambda_n \geq \beta R^*(\nabla \cL(\theta^*;Z^n))~
\label{eq:supp_cond1}
\eeq
where $R^*(\cdot)$ is the dual norm of $R(\cdot)$. Then the error vector $\hat{\Delta}_n = \hat{\theta}_{\lambda_n} - \theta^*$ belongs to the set:
\beq
E_r = E_r(\theta^*,\beta) = \left\{ \Delta \in \R^p ~ \left| ~ R(\theta^* + \Delta) \leq R(\theta^*) + \frac{1}{\beta} R(\Delta) \right. \right\}~.
\label{eq:supp_error1}
\eeq
}

\proof By the optimality of $\htn = \theta^* + \hdn$, we have
\beq
\cL(\theta^*+\hdn) + \lambda_n R(\theta^*+\hdn) - \left\{ \cL(\theta^*) + \lambda_n R(\theta^*) \right\} \leq 0~.
\eeq
Now, since $\cL$ is convex,
\beq
\cL(\theta^*+ \Delta) - \cL(\theta^*) \geq \langle \nabla \cL(\theta^*), \Delta \rangle \geq - | \langle \nabla \cL(\theta^*), \Delta \rangle |~.
\eeq
Further, by generalized Holder's inequality, we have
\beq
| \langle \nabla \cL(\theta^*), \Delta \rangle | \leq R^*(\nabla \cL(\theta^*)) R(\Delta) \leq \frac{\lambda_n}{\beta} R(\Delta)~,
\eeq
where we have used $\lambda_n \geq \beta R^*(\nabla \cL(\theta^*;Z^n))$. Hence, we have
\beq
\cL(\theta^*+\hdn) - \cL(\theta^*) \geq - \frac{\lambda_n}{\beta} R(\hdn)~.
\eeq
As a result,
\beq
\lambda_n \left\{ R(\theta^* + \hdn) - R(\theta^*) - \frac{1}{\beta} R(\hdn) \right\} \leq 0~.
\eeq
Noting that $\lambda_n > 0$ and rearranging completes the proof. \qed

\subsection{Relation between the Constrained and Regularized Error Cones}
In this section we show that the sizes of the regularized and constrained error sets are of the same order. Recall from~\cite{crpw12}, that the error set for the constrained setting for atomic norms is a cone given by:
\beq
C_c = C_c(\theta^*) = \text{cone}(E_c) = \text{cone}\left\{ \Delta \in \R^p ~ \left| ~ R(\theta^* + \Delta) \leq R(\theta^*) \right. \right\}~.
\eeq

The error set $E_r$ is given by:


\begin{align*}
 E_r = E_r(\theta^*,\beta) = \left\{ \Delta \in \R^p ~ \left| ~ R(\theta^* + \Delta) \leq R(\theta^*) + \frac{1}{\beta} R(\Delta) \right. \right\}~.
\end{align*}


Below we provide the proof of Theorem \ref{thm:relconsreg}.

\textbf{Theorem \ref{thm:relconsreg}}
\textit{
Let $A_c^{(\rho)} = E_c \cap \rho B_2^p$, $A_r^{(\rho)} = E_r \cap \rho B_2^p$, and $\bar{A}_c^{(\rho)} = C_c \cap \rho B_2^p$, where $\rho B_2^p = \{ u | \|u\|_2 \leq \rho\}$ is the $L_2$ ball of any radius $\rho > 0$. Then, for any $\beta > 1$ we have
\beq
w(A_c^{(\rho)}) \leq w(A_r^{(\rho)}) \leq \left ( 1 + \frac{2}{\beta - 1} \frac{\|\theta^*\|_2}{\rho} \right ) w(\bar{A}_c^{(\rho)}) ~,
\eeq
where $w(A)$ denotes the Gaussian width of any set $A$ given by: $w(A) = E_g\left[\underset{a\in A}{\sup}~\langle a,g \rangle\right]$, where $g$ is an isotropic Gaussian random vector, i.e., $g \sim N(0, \mathbb{I}_{p \times p})$.
}

\proof The first inequality simply follows from the fact that $E_c \subseteq E_r$ and Property 1 of Gaussian width. For the second part, from triangle inequality, we have
\beq
R(\Delta) \leq R(\theta^* + \Delta) + R(\theta^*)~.
\eeq
Then,
\begin{align*}
E_r(\theta^*,\beta) & = \left\{ \Delta \in \R^p \left| R(\theta^*+\Delta) \leq R(\theta^*) + \frac{1}{\beta}R(\Delta) \right. \right\} \\
& \subseteq \left\{ \Delta \in \R^p \left| R(\theta^*+\Delta) \leq R(\theta^*) + \frac{1}{\beta}R(\theta^* + \Delta) + \frac{1}{\beta} R(\theta^*) \right. \right\} \\
& = \left\{ \Delta \in \R^p \left| \left( 1 - \frac{1}{\beta} \right) R(\theta^*+\Delta) \leq \left(1 + \frac{1}{\beta} \right) R(\theta^*) \right. \right\} \\
& = \left\{ \Delta \in \R^p \left| R(\theta^*+\Delta) \leq \frac{\beta+1}{\beta-1} R(\theta^*) \right. \right\} = \tilde{E}_r(\theta^*,\beta)~.
\end{align*}
Let $\tilde{C}_r(\theta^{*}, \beta)$ denote the following set
\beq
\tilde{C}_r = \tilde{C}_r(\theta^*,\beta) = \text{cone}\left\{ \Delta - \frac{2}{\beta-1}\theta^{*} \left| \Delta \in \bar{E}_r \right. \right\} + \frac{2}{\beta-1} \theta^{*} ~.
\eeq
It follows naturally from the construction that $E_r \subseteq \tilde{C}_r$.

\begin{figure}
\begin{center}
\vspace*{-20mm}
\hspace*{-5mm}
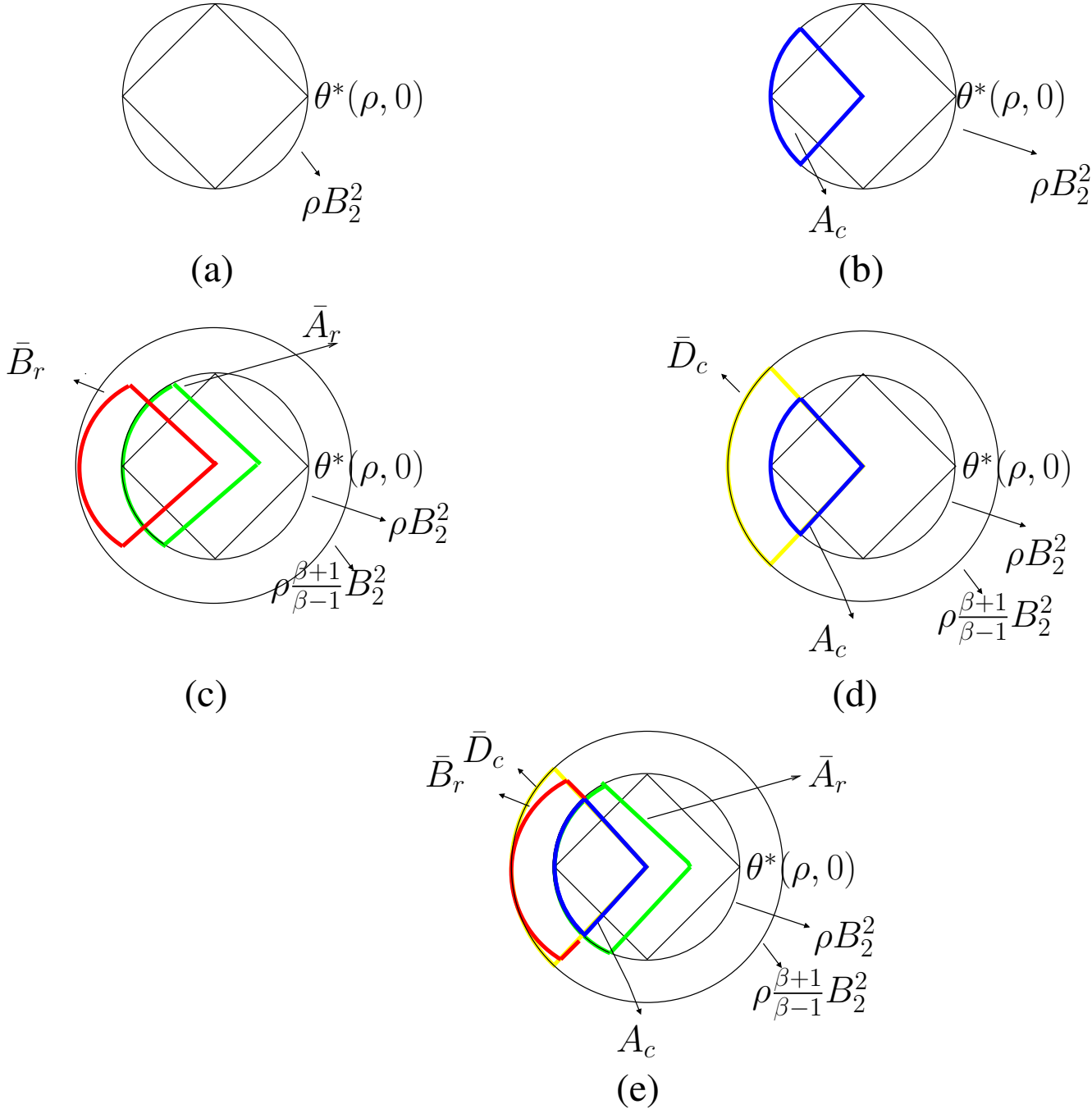
\caption{Error cone for $L_1$ norm in two dimensions: (a) The L1 norm ball in two dimensions; (b) The constrained error cone $A_c$; (c) The regularized error cone $\hat{A}_r$ and the shifted cone $\hat{B}_r$; (d) The constrained error cone $A_c$ and the shifted constrained error cone $\bar{D}_c$; (e) All error cones.}
\end{center}
\end{figure}

Let $\tilde{A}_r^{(\rho)} = \tilde{C}_r(\theta^*,\beta) \cap \rho B^{p}_{2}$. Since $E_r(\theta^*,\beta) \subseteq \tilde{C}_r(\theta^*,\beta)$, we have $w(A_r^{(\rho)}) \leq w(\tilde{A}_r^{(\rho)})$. We define two additional sets for our analysis:
\begin{align}
\tilde{B}_r^{(\rho)} & = \tilde{A}_r^{(\rho)} - \frac{2}{\beta - 1} \theta^* = \left\{ \Delta \in \R^p \left| \Delta + \frac{2}{\beta-1} \theta^* \in \tilde{A}_r^{(\rho)} \right. \right\}~,\\
\tilde{D}_c^{(\rho)} & = {C}_c(\theta^*, \beta) \cap \left ( \rho + \frac{2}{\beta - 1} \|\theta^*\|_2 \right ) B^p_2 = \left \{ \Delta \in \R^p \left | \Delta \in C_c, \|\Delta\|_2 \leq \left ( \rho + \frac{2}{\beta-1} \right ) \|\theta^*\|_2 \right. \right \}~.
\end{align}
Following Property 3 of Gaussian width, we have
\beq
w(\tilde{D}_c^{(\rho)}) = \left ( 1 + \frac{2}{\beta - 1} \frac{\|\theta^*\|_2}{\rho} \right ) w({\bar{A}}_c^{(\rho)})~.
\label{eq:tele1}
\eeq
Further, using Property 5 of Gaussian width, we have
\beq
w(\tilde{A}_r^{(\rho)}) = w(\tilde{B}_r^{(\rho)})~.
\label{eq:tele2}
\eeq
From the construction it is clear that $\tilde{B}_r^{(\rho)} \subset \tilde{D}_c^{(\rho)}$. Hence we have
\beq
w(\tilde{D}_c^{(\rho)}) \geq w(\tilde{B}_r^{(\rho)})
\eeq
Then, we have

\begin{align*}
w(\tilde{A}_r^{(\rho)}) = w(\tilde{B}_r^{(\rho)}) \leq w(\tilde{D}_c^{(\rho)}) = \left ( 1 + \frac{2}{\beta - 1} \frac{\|\theta^*\|_2}{\rho} \right ) w({\bar{A}}_c^{(\rho)})
\end{align*}

By noting that $w(A_r^{(\rho)}) \leq w(\tilde{A}_r^{(\rho)})$, we complete the proof. \qed

\subsection{Recovery Guarantees}
Lemma \ref{lem:recbnderror} and Theorem \ref{thm:recover} in the paper are results which establish recovery guarantees. The result in Lemma \ref{lem:recbnderror} depends on $\theta^*$, which is unknown. On the other hand Theorem \ref{thm:recover} gives the result in terms of quantities like $\lambda_n$ and the norm compatibility constant $\Psi(E_r) = \sup_{u \in E_r} \frac{R(u)}{\|u\|_2}$ which are easier to compute or bound. In this section we give proofs of Lemma \ref{lem:recbnderror} and Theorem \ref{thm:recover}.

\textbf{Lemma \ref{lem:recbnderror}}
\textit{
Assume that the RSC condition is satisfied in $E_r$ by the loss $\cL(\cdot)$ with parameter $\kappa$. With $\hdn = \htn - \theta^*$, for any norm $R(\cdot)$, we have
\beq
\| \hdn \|_2 \leq \frac{1}{\kappa} \| \nabla \cL(\theta^*) + \lambda_n \nabla R(\theta^*) \|_2~,
\eeq
where $\nabla R(\cdot)$ is any sub-gradient of the norm $R(\cdot)$.
}

\proof By the RSC property in $E_r$, for any $\Delta \in E_r$ we have
\beq
\cL(\theta^* + \Delta) - \cL(\theta^*) \geq \langle \nabla L(\theta^*), \Delta \rangle + \kappa \| \Delta \|_2^2~.
\label{eq:loss1}
\eeq
Also, recall that any norm is convex, since by triangle inequality, for $t \in [0,1]$, we have
\beq
R(t \theta_1 + (1-t) \theta_2) \leq R(t\theta_1) + R((1-t)\theta_2) = t R(\theta_1) + (1-t) R(\theta_2)~.
\eeq
As a result, for any sub-gradient $\nabla R(\theta)$ of $R(\theta)$, we have
\beq
R(\theta^* + \Delta) - R(\theta^*) \geq \langle \Delta, \nabla R(\theta^*) \rangle~.
\label{eq:reg1}
\eeq
Adding \myref{eq:loss1} and \myref{eq:reg1}, we get
\beq
\cL(\theta^* + \Delta) - \cL(\theta^*) + \lambda_n (R(\theta^* + \Delta) - R(\theta^*)) \geq \langle \nabla \cL(\theta^*) + \lambda_n \nabla R(\theta^*), \Delta \rangle + \kappa \| \Delta \|_2^2~
\label{eq:bnd1}
\eeq
Now, by Cauchy-Schwartz inequality, we have
\begin{align}
| \langle \nabla \cL(\theta^*) + \lambda_n \nabla R(\theta^*), \Delta \rangle | & \leq \| \nabla \cL(\theta^*) + \lambda_n \nabla R(\theta^*)\|_2 \|\Delta \|_2 \nonumber \\
\Rightarrow \qquad \langle \nabla \cL(\theta^*) + \lambda_n \nabla R(\theta^*), \Delta \rangle & \geq - \| \nabla \cL(\theta^*) + \lambda_n \nabla R(\theta^*)\|_2 \|\Delta \|_2~.
\label{eq:csbnd}
\end{align}
Using \myref{eq:csbnd} in \myref{eq:bnd1}, we have
\begin{align}
\cF(\Delta) & = \cL(\theta^* + \Delta) - \cL(\theta^*) + \lambda_n (R(\theta^* + \Delta) - R(\theta^*)) \nonumber \\
& \geq - \| \nabla \cL(\theta^*) + \lambda_n \nabla R(\theta^*)\|_2 \|\Delta \|_2 + \kappa \| \Delta \|_2^2 \nonumber \\
& = \kappa \| \Delta \|_2 \left\{ \| \Delta \|_2 - \frac{\| \nabla \cL(\theta^*) + \lambda_n \nabla R(\theta^*) \|_2}{\kappa} \right\}~.
\label{eq:genbnd}
\end{align}
Now, since $\cF(\hdn) \leq 0$, from \myref{eq:genbnd}, we have
\beq
\| \hdn \|_2 \leq \frac{\| \nabla \cL(\theta^*) + \lambda_n \nabla R(\theta^*) \|_2}{\kappa}~,
\eeq
which completes the proof. \qed

\textbf{Theorem \ref{thm:recover}}
\textit{
Assume that the RSC condition is satisfied in $E_r$ by the loss $\cL(\cdot)$ with parameter $\kappa$. With $\hdn = \htn - \theta^*$, for any norm $R(\cdot)$, we have
\beq
\| \hdn \|_2 \leq \frac{1+\beta}{\beta} \frac{\lambda_n}{\kappa}  \Psi(E_r)~.
\label{eq:supp_recover}
\eeq
}

\proof By the RSC property in $E_r$, we have for any $\Delta \in E_r$
\beq
\cL(\theta^* + \Delta) - \cL(\theta^*) \geq \langle \nabla L(\theta^*), \Delta \rangle + \kappa \| \Delta \|_2^2~.
\label{eq:loss2}
\eeq
By definition of a dual norm, we have
\beq
|\langle \nabla \cL(\theta^*), \Delta \rangle | \leq R^*(\nabla \cL(\theta^*)) R(\Delta)~.
\eeq
Further, by construction, $R^*(\nabla \cL(\theta^*)) \leq \frac{\lambda_n}{\beta}$, implying
\begin{align}
|\langle \nabla \cL(\theta^*), \Delta \rangle | & \leq \frac{\lambda_n}{\beta} R(\Delta)~\nonumber \\
\Rightarrow \qquad \langle \nabla \cL(\theta^*), \Delta \rangle & \geq - \frac{\lambda_n}{\beta} R(\Delta)~.
\label{eq:csbnd2}
\end{align}
Further, from triangle inequality, we have
\beq
R(\theta^* + \Delta) - R(\theta^*) \geq - R(\Delta)
\label{eq:normbnd2}
\eeq
Adding \myref{eq:csbnd2} and \myref{eq:normbnd2}, we have
\begin{align}
\cF(\Delta) & = \cL(\theta^* + \Delta) - \cL(\theta^*) + \lambda_n (R(\theta^* + \Delta) - R(\theta^*)) \geq - \frac{\lambda_n}{\beta} R(\Delta) + \kappa \| \Delta \|_2^2 - \lambda_n R(\Delta) \nonumber \\
&= \kappa \| \Delta \|_2^2 - \lambda_n \frac{1 + \beta}{\beta} R(\Delta)~.
\label{eq:genbnd2}
\end{align}
By definition of the norm compatibility constant $\psi_{E_r}$, we have $R(\Delta) \leq \| \Delta \|_2 \psi_{E_r}$ implying $-R(\Delta) \geq - \| \Delta \|_2 \psi_{E_r}$. Plugging the inequality back into \myref{eq:genbnd2}, we have
\begin{equation}
\cF(\Delta) \geq \kappa \| \Delta \|_2 \left\{ \| \Delta \|_2 - \frac{1+\beta}{\beta} \frac{\lambda_n}{\kappa} \psi_{E_r} \right\}~.
\end{equation}
Since $\cF(\hdn) \leq 0$, we have
\beq
\| \hdn \|_2 \leq \frac{1+\beta}{\beta} \frac{\lambda_n}{\kappa} \psi_{E_r}~,
\eeq
which completes the proof. \qed

\section{Bounds on the Regularization Parameter}
In this section, we prove Theorem \ref{theo:lambda_expect} and \ref{theo:lambda_gauss_bound} in Section \ref{sec:lambda} of the paper. The regularization parameter should satisfy the condition $\lambda_n \geq \beta R^*(\nabla \cL (\theta^*;Z^n))$. In Theorem \ref{theo:lambda_expect} we establish the upper bound on the expectation $E[R^*(\nabla \cL (\theta^*;Z^n))]$ in terms of the Gaussian width of the unit norm ball for least squares loss and Gaussian designs. In Theorem \ref{theo:lambda_gauss_bound} we show that $R^*(\nabla \cL (\theta^*;Z^n))$ concentrates sharply around its expectation.

\subsection{Proof of Theorem \ref{theo:lambda_expect}}
To prove Theorem \ref{theo:lambda_expect}, we first need the following theorem from generic chaining.
\begin{theo}
\label{theo:sub-gauss_expect}
Let $\Omega_R = \{ u : R(u) \leq 1 \}$ be the unit norm ball of $R(\cdot)$. Assuming $h$ is any centered sub-Gaussian random vector with $\vertiii{h}_{\psi_2} \leq \kappa$, then we have
\begin{align*}
E\left[\sup_{R(u)\leq 1} \langle h, u \rangle\right] \leq  \eta_0  \kappa w\left( \Omega_R \right)~,
\end{align*}
where $\eta_0$ is a universal constant.
\end{theo}
\proof The quantity $E[\sup_{R(u)\leq 1} \langle h, u \rangle]$ can be considered the ``sub-Gaussian width'' of $\Omega_R$, the unit norm ball, since it has the exact same form as the Gaussian width, with $h$ being a sub-Gaussian vector instead of a Gaussian vector. Next, we show that the sub-Gaussian width is always bounded by the Gaussian width times a factor proportional to $\kappa$.

Consider the sub-Gaussian process $Y = \{ Y_u \}, Y_u = \langle u, h \rangle$ indexed by $u \in \Omega_R$, the unit norm ball. Consider the Gaussian process $X = \{ X_u \}, X_u = \langle u, g \rangle$, where $g \sim N(0,\I)$, indexed by the same set, i.e., $u \in \Omega_R$, the unit norm ball. First, note that $|Y_u - Y_v| = |\langle h, u - v \rangle|$, so that by the concentration of sub-Gaussian random variable~\cite[Equation 5.10]{vers12}, we have
\beq
\begin{split}
P\left(|Y_u - Y_v | \geq \epsilon \right) \leq e \cdot \exp\left( - \frac{c \epsilon^2}{\kappa^2 \| u - v\|^2} \right)~,
\end{split}
\eeq
where $c > 0$ is an absolute constant. As a result, a direct application of the generic chaining argument for upper bounds on such empirical processes~\cite[Theorem 2.1.5]{tala05} gives
\beq
E\left[ \sup_{u,v} | Y_u - Y_v | \right] \leq \eta_1 E\left[ \sup_u X_u \right] = \eta_1 w(\Omega_R)~,
\eeq
where $\eta_1$ is an absolute constant. Further, since $\{Y_u\}$ is a symmetric process, from~\cite[Lemma 1.2.8]{tala05}, we have
\beq
E\left[ \sup_{u,v} | Y_u - Y_v | \right] = 2 E\left[ \sup_u Y_u \right]~.
\eeq
As a result, with $\eta_0 = \eta_1/2$, we have
\beq
E\left[\sup_{R(u)\leq 1} \langle h, u \rangle\right] = E\left[ \sup_u Y_u \right] \leq \eta_0 w(\Omega_R)~.
\eeq
That completes the proof. \qed

Now we turn to the proof of Theorem \ref{theo:lambda_expect}. \\
\textbf{Theorem \ref{theo:lambda_expect}}
\textit{
Let $\Omega_R = \{ u: R(u) \leq 1\}$, and $\cal L$ be the squared loss. For sub-Gaussian design $X$ and noise $\omega$, we have
\beq
E\left[R^*(\nabla {\cal L}(\theta^*;Z^n))\right] \leq \eta \xi \cdot \frac{\kappa w(\Omega_R)}{\sqrt{n}} ~,
\vspace*{-2mm}
\eeq
where the expectation is taken over both $X$ and $\omega$. The constant $\xi$ is given by
\begin{align*}
\xi = \left \{
             \begin{array}{lll}
              1 \ \ \ &\text{if $X$ is isotropic} \\
              \sqrt{\Lambda_{\max}(\Sigma)} \ \ \ &\text{if $X$ is anisotropic} ~. \\
             \end{array}  \right.
\end{align*}
}

\proof For least squares loss, we first note that
\begin{equation}
\label{eq:decouple}
\begin{split}
E\left[R^*(\nabla {\cal L}(\theta^*;Z^n))\right] &= E\left[R^*(\frac{1}{n} X^T \omega)\right] = E\left[\sup_{R(u)\leq 1} \big \langle \frac{1}{n} X^T \omega, u \big \rangle \right]  \\
&= E\left[\frac{1}{n} \|\omega\|_2 \cdot E\left[\sup_{R(u)\leq 1} \Big \langle X^T \frac{\omega}{\|\omega\|_2}, u \Big \rangle \Big | \omega \right] \right] \\
&\leq E\left[\frac{1}{n} \|\omega\|_2 \cdot \sup_{v \in \s^{n-1}} E\left[\sup_{R(u)\leq 1} \Big \langle X^T v, u \Big \rangle \right] \right] \\
&= \frac{1}{n} E\left[ \|\omega\|_2 \right] \cdot \sup_{v \in \s^{n-1}} E\left[\sup_{R(u)\leq 1} \Big \langle X^T v, u \Big \rangle \right] ~.
\end{split}
\end{equation}
$E[\|\omega\|_2]$ is the expected length of a centered sub-Gaussian random vector, which can be easily bounded using Jensen's inequality,
\beq
\label{eq:length_bound}
 E [\|\omega\|_2 ] < \sqrt{E [\|\omega\|_2^2 ]} = \sqrt{n} ~.
\eeq
Then we focus on $E\left[\underset{R(u)\leq 1}{\sup} \langle X^T v, u \rangle\right]$ for any fixed $v \in \s^{n-1}$. Let $h = X^T v$ be a random vector, and consider the random variable $\langle h, z \rangle$ for any fixed $z \in \s^{p-1}$. Note that
\begin{align*}
\langle h, z \rangle = \langle v,  X z \rangle ~.
\end{align*}
\textbf{Case 1.} If $X$ is independent isotropic, then $Xz$ has i.i.d. centered sub-Gaussian entries with $\psi_2$ norm at most $\kappa$. By Lemma \ref{lem:rotinvsubg}, we know that $\langle v, Xz \rangle$ is sub-Gaussian with $\vertiii{\langle v, Xz\rangle}_{\psi_2} \leq C\kappa$, where $C$ is an absolute constant. Hence $h$ is a sub-Gaussian random vector with $\vertiii{h}_{\psi_2} \leq C\kappa$. Using Theorem \ref{theo:sub-gauss_expect}, we conclude that for any $v \in \s^{n-1}$
\beq
\label{eq:chaining_bound_iso}
E\left[\underset{R(u)\leq 1}{\sup} \Big \langle X^T v, u \Big \rangle\right] \leq \eta_0 C \kappa \cdot w(\Omega_R) ~.
\eeq
\textbf{Case 2.} If $X$ is independent anisotropic, then $Xz$ has i.i.d. centered sub-Gaussian entries with $\psi_2$ norm at most $\kappa \sqrt{\Lambda_{\max}(\Sigma)} $, where $\Lambda_{\max}(\Sigma)$ is the largest eigenvalue of $\Sigma$. By the same argument as Case 1, we have
\beq
\label{eq:chaining_bound_aniso}
E\left[\underset{R(u)\leq 1}{\sup} \Big \langle X^T v, u \Big \rangle\right] \leq \eta_0 C \kappa \sqrt{\Lambda_{\max}(\Sigma)} \cdot  w(\Omega_R) ~.
\eeq
Letting $\eta = \eta_0 C$ and combining \eqref{eq:chaining_bound_iso}, \eqref{eq:length_bound} and \eqref{eq:chaining_bound_aniso}, we complete the proof. \qed

\subsection{Proof of Theorem \ref{theo:lambda_gauss_bound}}
To prove Theorem \ref{theo:lambda_gauss_bound}, we also need the following result from generic chaining.
\begin{theo}
\label{theo:sub-gauss_conc}
Let $\Omega_R = \{ u : R(u) \leq 1 \}$ be the unit norm ball of $R(\cdot)$. Assuming $h$ is any centered sub-Gaussian random vector with $\vertiii{h}_{\psi_2} \leq \kappa$, then we have for any $\tau > 0$,
\beq
\label{eq:subgauss_conc}
P\left( \sup_{R(u) \leq 1} \langle h, u \rangle \geq  \nu_0 \kappa w(\Omega_R) + \tau \right) \leq \nu_1 \exp \left(-\left(\frac{\tau}{\nu_2 \kappa \phi}\right)^2\right) ~,
\eeq
where $\nu_0$, $\nu_1$ and $\nu_2$ are universal constants, and $\phi = \sup_{R(u) \leq 1} \|u\|_2$.
\end{theo}
\proof Consider the sub-Gaussian process $Y = \{ Y_u \}, Y_u = \langle h, u \rangle$ indexed by $u \in \Omega_R$. By the same argument in the proof of Theorem \ref{theo:sub-gauss_expect}, we have
\beq
\begin{split}
P\left(|Y_u - Y_v | \geq \epsilon \right) \leq e \cdot \exp\left( - \frac{c \epsilon^2}{\kappa^2 \| u - v\|^2} \right)~,
\end{split}
\eeq
and
\beq
E\left[ \sup_{u,v} | Y_u - Y_v | \right] = 2 E\left[ \sup_u Y_u \right]~.
\eeq
Then a direct application of ~\cite[Theorem 2.1.5]{tala05} and ~\cite[Theorem 2.2.27]{tala14} gives us \eqref{eq:subgauss_conc}.  \qed

\textbf{Theorem \ref{theo:lambda_gauss_bound}}
\textit{
Let design $X$ and noise $\omega$ be sub-Gaussian, and $\cal L$ be squared loss. Define $\phi = \sup_{R(u) \leq 1} \|u\|_2$, then for any $\tau > 0$, with probability at least $1- c_1 \exp\left(- \min\left((\frac{\tau}{c_2 \xi \kappa \phi})^2, c_0 n\right)\right)$, we have
\beq
R^*\left(\nabla {\cal L}(\theta^*;Z^n)\right) \leq  \sqrt{\frac{2 K^2 + 1}{n}} \left(c \xi \kappa \cdot w(\Omega_R) + \tau\right) ~,
\label{eq:gausDualBound}
\eeq
where $c$, $c_0$, $c_1$ and $c_2$ are all absolute constants, and $\xi$ is the same as in Theorem \ref{theo:lambda_expect}. \\}
\proof We only show the case for isotropic $X$, where $\xi = 1$. Note that
\beq
\begin{split}
&\ \ \ \ \ \ P\left(R^*\left(\nabla {\cal L}(\theta^*;Z^n)\right) \geq \sqrt{\frac{2 K^2 + 1}{n}} \left(c \xi \kappa  \cdot w(\Omega_R) + \tau\right)\right) \\
&= P\left(\|\omega\|_2 \cdot R^*\left( \frac{X^T\omega}{\|\omega\|_2} \right) \geq \sqrt{(2 K^2 + 1)n} \left( c \kappa  \cdot w(\Omega_R) + \tau\right)\right) \\
&\leq P\left( \|\omega\|_2 > \sqrt{(2 K^2 + 1)n} \right) + \sup_{v \in \s^{n-1}} P\left(R^*\left( X^T v \right) \geq c \kappa \cdot w(\Omega_R) + \tau \right) ~,
\end{split}
\eeq
where the last inequality uses the union bound. We first prove the bound for $\|\omega\|_2$.
Since $\omega$ consists of i.i.d. centered unit-variance sub-Gaussian elements with $\vertiii{ \omega_i }_{\psi_2} < K$, $\omega_i^2$ is sub-exponential with $\vertiii{ \omega_i }_{\psi_1} < 2K^2$. By applying Bernstein's inequality to $\|\omega\|_2^2 = \sum_{i=1}^{n} \omega_i^2$, we obtain
\begin{align*}
P\left(\Big |\|\omega\|_2^2 - E[\|\omega\|_2^2 ] \Big | \geq \tau\right) \leq 2 \exp \left[-c_0 \min\left(\frac{\tau^2}{4 K^4 n}, \frac{\tau}{2K^2}\right)\right] ~,
\end{align*}
where $c_0$ is an absolute constant. Setting $\tau = 2 K^2 n$ and using \eqref{eq:length_bound}, we get
\begin{align}
\label{eq:bound_length_subgauss}
P\left(\|\omega\|_2 \geq \sqrt{(2K^2 + 1)n} \right) \leq 2 \exp (-c_0 n)
\end{align}
Next we bound $R^*(X^T v)$ for any $v \in \s^{n-1}$. Given any fixed $v \in \s^{n-1}$, we note that
\begin{align*}
R^*(X^T v) = \sup_{R(u) \leq 1} \langle X^T v, u \rangle ~,
\end{align*}
and $X^T v$ is a sub-Gaussian random vector with $\vertiii{X^T v}_{\psi_2} \leq C \kappa$ as shown in the proof of Theorem \ref{theo:lambda_expect}. Using Theorem \ref{theo:sub-gauss_conc}, we have
\beq
\label{eq:bound_sup_subgauss}
P\left( R^*(X^T v) \geq  \nu_0 C \kappa  \cdot w(\Omega_R) + \tau \right) \leq \nu_1 \exp \left(-\left(\frac{\tau}{\nu_2 C \kappa \phi}\right)^2\right) ~.
\eeq
Letting $c = \nu_0 C$, $c_1 = \nu_1 + 2$ and $c_2 = \nu_2 C$, and combining \eqref{eq:bound_length_subgauss} and \eqref{eq:bound_sup_subgauss}, we complete the proof. \qed

\subsection{Proof of Lemma~\ref{lem:ballWidthBound}}
\textbf{Lemma \ref{lem:ballWidthBound}}
\textit{
Let $\Omega_R = \{ u : R(u) \leq 1 \}$ be the unit norm ball and $\Theta_R = \{ u : R(u) = 1 \}$ be the boundary. For any $\tilde{\theta} \in \Theta_R$ define $\rho(\tilde{\theta}) = \sup_{\theta:R(\theta) \leq 1} \|\theta - \tilde{\theta}\|_2$ is the diameter of $\Omega_R$ measured with respect to $\tilde{\theta}$. If $G(\tilde{\theta}) = \cone(\Omega_R - \tilde{\theta}) \cap \rho(\tilde{\theta}) B_2^p$, i.e., the cone of $(\Omega_R - \tilde{\theta})$ intersecting the ball of radius $\rho(\tilde{\theta})$. Then
\beq
w(\Omega_R) \leq \inf_{\tilde{\theta} \in \Theta_R} ~w(G(\tilde{\theta}))
\eeq}
\proof For any $\tilde{\theta} \in \Theta_R$, consider the set $F_R(\tilde{\theta}) = \Omega_R - \tilde{\theta} = \{ u: R(u + \tilde{\theta}) \leq 1 \}$.
Since Gaussian width is translation invariant, the Gaussian width of $\Omega_R$ and $F_R$ are the same, i.e., $w(\Omega_R) = w(F_R(\tilde{\theta}))$. Since, $\rho(\tilde{\theta}) = \sup_{\theta: R(\theta) \leq 1} \| \theta - \tilde{\theta} \|_2$ is the diameter of $\Omega_R$ as well as $F_R(\tilde{\theta})$, a ball of radius $\rho(\tilde{\theta})$ will include $F_R(\tilde{\theta})$, so that $F_R(\tilde{\theta}) \subseteq \rho(\tilde{\theta}) B_2^p$. Further, by definition, $F_R(\tilde{\theta}) \subseteq \text{cone}(F_R(\tilde{\theta})) = \text{cone}(\Omega_R - \tilde{\theta})$. Let $G(\tilde{\theta}) = \text{cone}(\Omega_R - \tilde{\theta}) \cap \rho(\tilde{\theta}) B_2^p$. By construction, $F_R(\tilde{\theta}) \subseteq G(\tilde{\theta})$. Then,
\begin{align*}
w(\Omega_R) = w(F_R(\tilde{\theta})) \leq w(G(\tilde{\theta}))~.
\end{align*}
Noting the analysis holds for any $\tilde{\theta} \in \Theta_R$, completes the proof. \qed

\section{Restricted Eigenvalue Conditions: Sub-Gaussian Designs}
We focus on results in Section \ref{sec:re}. In particular we consider RE conditions for sub-Gaussian design matrices for three different cases: (i) the design matrix has independent sub-Gaussian rows $X_i$ with $\vertiii{X_i}_{\psi_2} \leq \kappa$, (ii) the design matrix has independent rows with subGaussian elements $x_{ij}$ so that  $\vertiii{x_{ij}}_{\psi_2} \leq \kappa$ and the columns are correlated, and (iii) the columns are independent subGaussian but the rows are correlated, i.e., correlated samples. One can view (ii) as a special case of (i), but we highlight this special case because of its practical importance and past literature on RE conditions for anisotropic subGaussian designs~\cite{rawy10,ruzh13}.

%

Our results simply use a general treatment developed in~\cite{mept07}, building on~\cite{klme05}, based on Talagrand's generic chaining~\cite{tala05,tala14}. We specifically focus on results in~\cite{mept07} which provide uniform bounds on the supremum of certain empirical processes. RE results for the specific cases of interest in the current paper will then be established by suitable choices of these empirical processes. The results in~\cite{mept07}, and more generally in generic chaining~\cite{tala05,tala14}, are based on certain $\gamma$-functionals which we briefly introduce below.

Consider a metric space $(T,d)$ and for a finite set ${\cal A} \subset T$, let $|{\cal A}|$ denote its cardinality. An {\em admissible sequence} is an increasing sequence of subsets $\{ {\cal A}_n, n \geq 0\}$ of $T$, such that $|{\cal A}_0|=1$ and for $n \geq 1$, $|{\cal A}_n| = 2^{2^n}$. Given $\alpha > 0$, we define the $\gamma_{\alpha}$-functional as
\beq
\gamma_{\alpha}(T,d) = \inf \sup_{t \in T} \sum_{n=0}^{\infty} \text{Diam}(A_n(t))~,
\eeq
where $A_n(t)$ is the unique element of ${\cal A}_n$ that contains $t$, $\text{Diam}(A_n(t))$ is the diameter of $A_n$ according to $d$, and the infimum is over all admissible sequences of $T$.
To get the desired RIP results in terms of Gaussian widths, we start with the following key result, originally \cite[Theorem D]{mept07}.
\begin{theo}[Mendelson, Pajor, Tomczak-Jaegermann~\cite{mept07}]
There exist absolute constants $c_1$, $c_2$, $c_3$ for which the following
holds. Let $(\Omega,\mu)$ be a probability space, set $F$ be a subset of the unit
sphere of $L_2(\mu)$, i.e., $F \subseteq S_{L_2} = \{ f : \vertiii{f}_{L_2} = 1\}$, and assume that $ \sup_{f \in F}~\vertiii{f}_{\psi_2} \leq \kappa$. Then, for any $\theta > 0$ and $n \geq 1$ satisfying
\beq
c_1 \kappa \gamma_2(F, \vertiii{\cdot}_{\psi_2}) \leq \theta \sqrt{n}~,
\eeq
with probability at least $1- \exp(-c_2 \theta^2 n/\kappa^4)$,
\beq
\sup_{f \in F}~\left| \frac{1}{n} \sum_{i=1}^n f^2(X_i) - E\left[f^2\right] \right| \leq \theta~.
\eeq
Further, if $F$ is symmetric, then
\beq
E\left[ \sup_{f \in F} ~\left| \frac{1}{n} \sum_{i=1}^n f^2(X_i) - E\left[f^2\right] \right| \right] \leq c_3 \max \left\{ 2\kappa \frac{\gamma_2(F, \vertiii{\cdot}_{\psi_2})}{\sqrt{n}}, \frac{\gamma_2^2(F, \vertiii{\cdot}_{\psi_2})}{n} \right\}
\eeq
\label{thm:subGREmain}
\end{theo}


We use the above result and related arguments to establish RIP conditions for the cases of interest.

\subsection{Isotropic Sub-Gaussian Designs}
We consider the case where the design matrix $X \in \R^{n \times p}$ has independent subGaussian rows where each row satisfies $\vertiii{X_i}_{\psi_2} \leq \kappa$ and $E[X_i X_i^T] = \I_{p \times p}$. Thus, the measure $\mu$ from which the rows $X_i$ are sampled independently is an isotropic sub-Gaussian measure.
\begin{theo}
Let $X$ be a design matrix with independent isotropic subGaussian rows, i.e., $\vertiii{X_i}_{\psi_2} \leq \kappa$ and $E[X_i X_i^T] = \I$. Then, for absolute constants $\eta, c > 0$, with probability at least $(1-2 \exp(-\eta w^2(A)))$, we have
\beq
\sup_{u \in A} \left| \frac{1}{n} ||Xu||^2 - 1 \right| ~=~\sup_{u \in A} \left| \frac{1}{n} \sum_{i=1}^n \langle X_i, u \rangle^2 - 1 \right| ~\leq ~c\frac{w(A)}{\sqrt{n}}~,
\eeq
or, equivalently,
\beq
1 - c\frac{w(A)}{\sqrt{n}} ~~\leq ~~\inf_{u \in A}~ \frac{1}{n} ||Xu||^2 ~~\leq ~~\sup_{u \in A} ~\frac{1}{n} ||Xu||^2  ~~\leq ~~1 + c\frac{w(A)}{\sqrt{n}}~.
\eeq
\label{thm:isg-gc}
\end{theo}
\proof The result essentially follows from an application of Theorem~\ref{thm:subGREmain}.
For convenience of notation, let $X_0$ be i.i.d.~as the rows $X_i, i=1,\ldots,n$, thus distributed following $\mu$.
To apply Theorem~\ref{thm:subGREmain}, we choose any $A \subseteq S^{p-1}$ consider the following class of functions: $F = \{ \langle \cdot, u \rangle : u \in A \}$. Then, $f(X_0) = \langle X_0, u \rangle$ and $F$ is a subset of the unit sphere, i.e., $F \subseteq S_{L_2}$, since $\vertiii{f}_{L_2} = E[u^T X_0^T X_0 u] = \| u \|^2 = 1$. Further, $ \sup_{f \in F}~\vertiii{f}_{\psi_2} = \sup_{u \in A} \vertiii{\langle X_0, u \rangle}_{\psi_2} \leq \vertiii{X_0}_{\psi_2} \leq \kappa/2$.

Next, we show that for the current setting, the $\gamma_2$-functional can be upper bounded by $w(A)$, the Gaussian width of $A$. Since $\mu$ is isotropic subGaussian with $\psi_2$-norm bounded by $\kappa$, we have
\beq
\gamma_2(F \cap S_{L_2}, \vertiii{\cdot}_{\psi_2}) ~\leq ~\kappa \gamma_2(F \cap S_{L_2}, \vertiii{\cdot}_{L_2}) ~\leq ~ \kappa c_4 w(A)~,
\eeq
where the last inequality follows from generic chaining, in particular \cite[Theorem 2.1.1]{tala05}, for an absolute constant $c_4 > 0$.

In the context of Theorem~\ref{thm:subGREmain}, we choose
\[
\theta = c_1 c_4 \kappa^2 \frac{w(A)}{\sqrt{n}} \geq c_1 \kappa \frac{ \gamma_2(F \cap S_{L_2}, \vertiii{\cdot}_{\psi_2})}{\sqrt{n}}~,
\]
so that the condition on $\theta$ is satisfied. With this choice of $\theta$, we have
\[
\theta^2 n/\kappa^4 = c_1^2 c_4^2 w^2(A)~.
\]
Then, from Theorem~\ref{thm:subGREmain}, it follows that with probability at least $1 - \exp(-\eta w^2(A))$, we have
\beq
\sup_{u \in A} \left| \frac{1}{n} \sum_{i=1}^n \langle X_i, u \rangle^2 - 1 \right|~ \leq c \frac{w(A)}{\sqrt{n}}~,
\eeq
where $\eta = c_2 c_1^2 c_4^2$ and $c = c_1 c_2 \kappa^2$ are absolute constants. As a result, we have
\[
\sup_{u \in A} \left( \frac{1}{n} \sum_{i=1}^n \langle X_i, u \rangle^2 - 1 \right) ~ \leq ~ c \frac{w(A)}{\sqrt{n}}~, \qquad \text{and} \qquad
\sup_{u \in A} \left( 1 - \frac{1}{n} \sum_{i=1}^n \langle X_i, u \rangle^2 \right) ~ \leq ~ c  \frac{w(A)}{\sqrt{n}}~,
\]
yielding
\beq
1 - c\frac{w(A)}{\sqrt{n}} ~~\leq ~~\inf_{u \in A} ~\frac{1}{n} ||Xu||^2 ~~\leq ~~\sup_{u \in A} ~\frac{1}{n} ||Xu||^2  ~~\leq ~~1 + c\frac{w(A)}{\sqrt{n}}~.
\eeq
That completes the proof. \qed

\subsection{Anisotropic Sub-Gaussian Designs}
We now consider the case where the design matrix $X \in \R^{n \times p}$ has independent rows, and each row $X_i$ is anisotropic subGaussian with $E[X_i^T X_i] = \Sigma$. Further, we assume that corresponding isotropic random vector $\tilde{X}_i = X_i \Sigma^{-1/2}$ satisfies $\vertiii{\tilde{X}_i}_{\psi_2} \leq \kappa$. A simple special case of such an anisotropic subGaussian design is when $X_i \sim N(0,\Sigma)$, where $\tilde{X}_i = X_i \Sigma^{-1/2} \sim N(0,\I)$ so that $\vertiii{\tilde{X}_i}_{\psi_2} = 1$. The result below characterizes RIP-style property of any such anisotropic subGaussian designs.
\begin{theo}
Let $X$ be a design matrix with independent anisotropic subGaussian rows, i.e., $E[X_i^T X_i] = \Sigma$ and $\vertiii{X_i \Sigma^{-1/2}}_{\psi_2} \leq \kappa$. Then, for absolute constants $\eta, c > 0$, with probability at least $(1-2 \exp(-\eta w^2(A)))$, we have
\beq
\sup_{u \in A} \left| \frac{1}{n} \frac{1}{u^T \Sigma u} ||Xu||^2 - 1 \right| ~=~\sup_{u \in A} \left| \frac{1}{n} \frac{1}{u^T \Sigma u} \sum_{i=1}^n \langle X_i, u \rangle^2 - 1 \right| ~\leq ~c\frac{w(A)}{\sqrt{n}}~.
\eeq
Further,
\beq
\lambda_{\min}(\Sigma|A)\left( 1 - c\frac{w(A)}{\sqrt{n}} \right) ~~\leq ~~\inf_{u \in A}~ \frac{1}{n} ||Xu||^2 ~~\leq ~~\sup_{u \in A} ~\frac{1}{n} ||Xu||^2  ~~\leq ~~\lambda_{\max}(\Sigma|A) \left(1 + c\frac{w(A)}{\sqrt{n}} \right)~,
\eeq
where
\beq
\lambda_{\min}(\Sigma|A) = \inf_{u \in A}~u^T \Sigma u~, \qquad \text{and} \qquad \lambda_{\max}(\Sigma|A) = \sup_{u \in A}~u^T \Sigma u~
\eeq
are the restricted minimum and maximum eigenvalues of $\Sigma$ restricted to $A \subseteq S^{p-1}$.
\label{thm:iasg-gc}
\end{theo}

\proof The result also follows from an application of Theorem~\ref{thm:subGREmain}.
For convenience of notation, let $X_0$ be i.i.d.~as the rows $X_i, i=1,\ldots,n$, thus distributed following $\mu$.
To apply Theorem~\ref{thm:subGREmain}, we choose any $A \subseteq S^{p-1}$ consider the following class of functions:
\beq
F = \{ f_u, u \in A : f_u(\cdot) = \frac{1}{\sqrt{u^T \Sigma u}}~\langle \cdot, u \rangle \}~.
\eeq
Then, $f_u(X_0) = \frac{1}{\sqrt{u^T \Sigma u}}~\langle X_0, u \rangle$ and $F$ is a subset of the unit sphere, i.e., $F \subseteq S_{L_2}$, since for $f_u \in F$
\[
\vertiii{f_u}_{L_2}^2 = \frac{1}{u^T \Sigma u} E[u^T X_0^T X_0 u] =  1~.
\]
Next, we focus on getting an upper bound on $\sup_{f_u \in F}~\vertiii{f_u}_{\psi_2} = \sup_{u \in A} \vertiii{\frac{1}{\sqrt{u^T \Sigma u}} \langle X_0, u \rangle}_{\psi_2}$. Let $\tilde{X}_0 = X_0 \Sigma^{-1/2}$ so that $\tilde{X}_0$ is a isotropic vector with $\vertiii{ \tilde{X}_0 }_{\psi_2} \leq \kappa$. Noting that
\[
\langle X_0, u \rangle = \langle \tilde{X}_0, \Sigma^{1/2} u \rangle = \sqrt{ u^T \Sigma u } \langle \tilde{X}_0, \frac{\Sigma^{1/2} u}{\| \Sigma^{1/2} u \|_2} \rangle~,
\]
we have
\[
\sup_u \vertiii{ f_u }_{\psi_2}  =  \sup_{u \in A} \vertiii{\frac{1}{\sqrt{u^T \Sigma u}} \langle X_0, u \rangle}_{\psi_2}
 =  \sup_{u \in A}~ \vertiii{ \langle \tilde{X}_0,  \frac{\Sigma^{1/2} u}{\| \Sigma^{1/2} u \|_2} \rangle }_{\psi_2} \leq \vertiii{ \tilde{X}_0 }_{\psi_2} \leq \kappa~.
\]
As a result, we have
\beq
\gamma_2(F \cap S_{L_2}, \vertiii{\cdot}_{\psi_2}) ~~\leq ~~\kappa~ \gamma_2(F \cap S_{L_2}, \vertiii{\cdot}_{L_2}) ~~\leq ~~ \kappa c_4 w(A)~,
\eeq
where the last inequality follows from \cite[Theorem 2.1.1]{tala05}, for an absolute constant $c_4 > 0$.

In the context of Theorem~\ref{thm:subGREmain}, we choose
\[
\theta = c_1 c_4 \kappa^2 \frac{w(A)}{\sqrt{n}} \geq c_1 \kappa \frac{ \gamma_2(F \cap S_{L_2}, \vertiii{\cdot}_{\psi_2})}{\sqrt{n}}~,
\]
so that the lower bound condition on $\theta$ is satisfied. With this choice of $\theta$, we have
\[
\theta^2 n/\kappa^4 = c_1^2 c_4^2 w^2(A)~.
\]
Then, from Theorem~\ref{thm:subGREmain}, it follows that with probability at least $1 - \exp(-\eta w^2(A))$, we have
\beq
\sup_{u \in A} \left| \frac{1}{n}  \frac{1}{u^T \Sigma u} \sum_{i=1}^n \langle X_i, u \rangle^2 - 1 \right|~ \leq c \frac{w(A)}{\sqrt{n}}~,
\eeq
where $\eta = c_2 c_1^2 c_4^2$ and $c = c_1 c_2 \kappa^2$ are absolute constants. As a result, we have
\begin{align}
\sup_{u \in A} \left( \frac{1}{n} \frac{1}{u^T \Sigma u} \sum_{i=1}^n \langle X_i, u \rangle^2 - 1 \right) & \leq ~ c \frac{w(A)}{\sqrt{n}}~, \label{eq:rip1-aniso} \\
\sup_{u \in A} \left( 1 - \frac{1}{n} \frac{1}{u^T \Sigma u} \sum_{i=1}^n \langle X_i, u \rangle^2 \right) & \leq ~ c  \frac{w(A)}{\sqrt{n}}~. \label{eq:rip2-aniso}
\end{align}
From \myref{eq:rip1-aniso}, we have
\[
\frac{1}{\lambda_{\max}(\Sigma|A)} \sup_{u \in A} \frac{1}{n} \| X u \|^2 ~ \leq ~ \sup_{u \in A} \frac{1}{n} \frac{1}{u^T \Sigma u} \| X u \|^2 ~\leq ~1 + c \frac{w(A)}{\sqrt{n}}~,
\]
so that
\beq
\sup_{u \in A}~\frac{1}{n} \| X u \|^2 \leq \lambda_{\max}(\Sigma|A) + c \lambda_{\max}(\Sigma|A) \frac{w(A)}{\sqrt{n}}~.
\label{eq:right-aniso}
\eeq
Similarly, from \myref{eq:rip2-aniso}, we have
\[
1 - c \frac{w(A)}{\sqrt{n}} ~\leq ~\inf_{u \in A} \frac{1}{n} \frac{1}{u^T \Sigma u} \| X u \|^2 ~\leq~ \frac{1}{\lambda_{\min}(\Sigma|A)} \inf_{u \in A} \frac{1}{n} \| X u \|^2~,
\]
implying
\beq
\lambda_{\min}(\Sigma|A) - c \lambda_{\min}(\Sigma|A) \frac{w(A)}{\sqrt{n}} \leq \inf_{u \in A} \frac{1}{n} \| X u \|^2~.
\label{eq:left-aniso}
\eeq
Putting \myref{eq:right-aniso} and \myref{eq:left-aniso} together completes the proof. \qed

\section{Generalized Linear Models: Restricted Strong Convexity}

We establish bounds on the regularization parameter and RSC condition for GLMs as discussed in Section~\ref{sec:glm}, along with a few specific examples.










\subsection{Generalized Linear Models}
Loss functions for GLMs are derived as maximum likelihood estimators for the family of exponential distributions. The canonical density function of exponential family distributions is given by~\cite{barn78,brow86,wajo08}:
\beq
P(y|\eta) \propto \exp \{ \eta y - \varphi(\eta) \} ~,
\eeq
where $\eta$ is the natural parameter and has a one-to-one function mapping with the mean parameter $\mu = E[y]$ of the distribution, $\varphi(\eta)$ is the log-partition function which ensures that $P(y|\eta)$ remains a probability distribution. The gradient of the log-partition function is the response function, i.e., $g(\cdot) = \varphi'(\cdot)$, which is monotonic by construction. The inverse of the response function is the so-called link function $h(\cdot) = g^{-1}(\cdot)$. The mean of the distribution can be obtained from the gradient of the log-partition function at the natural parameter, i.e.,
\beq
\mu = \varphi^{'}(\eta) = g(\eta) ~.
\eeq
The interested reader can study details and additional properties of exponential families from the existing literature~\cite{brow86,barn78,wajo08}.
Examples of distributions from the exponential family include the Gaussian, multinomial, exponential, Dirichlet, Poisson, Gamma, etc.

GLMs are obtained from conditional exponential family distributions by assuming a suitable parametric form of the natural parameter $\eta$ in terms of $X$ and $\theta^*$, in particular $\eta_i = \langle  X_i, \theta^* \rangle$. Then, the conditional distribution is given by
\beq
P(y_i|X_i,\theta^*) \propto \exp \{ \eta_i y_i - \varphi(\eta_i) \} = \exp \{ \langle X_i,\theta^* \rangle y_i - \varphi(\langle X_i,\theta^* \rangle) \} ~.
\eeq
The loss function for GLMs simply consider the negative log likelihood of such conditional exponential family forms. Assuming samples to be independent, we have
\beq
\cL(\theta;Z^n) = - \frac{1}{n} \sum_{i=1}^n \left\{ \eta_i y_i - \varphi(\eta_i) \right\} = - \frac{1}{n} \sum_{i=1}^n \left\{ \langle X_i y_i, \theta \rangle - \varphi(\langle X_i, \theta \rangle) \right\} ~.
\eeq
Using chain rule, the first derivative of the loss function evaluated at $\theta^*$ is
\begin{align*}
\nabla_{\theta} \cL(\theta^{*};Z^{n}) &= -\frac{1}{n} \sum_{i=1}^{n} y_{i}X_{i} + \frac{1}{n} \sum_{i=1}^{n} X_{i} \frac{\partial\varphi(\langle \theta^*, X_{i} \rangle)}{\partial \eta_i} = \frac{1}{n} \sum_{i=1}^n X_i(E[y_i|X_i] - y_i) = \frac{1}{n}X^{T} \omega ~,
\end{align*}
where each element of $\omega \in R^n$ is given as $\omega_i = E(y|X_i) - y_i$. Next we look at some specific examples of exponential families and corresponding GLMs.


\vspace*{-2mm}
{\leftmargini=2.6ex
\begin{enumerate}
\item Gaussian distribution: If the variance of the Gaussian distribution $P(y|X_i)$ is assumed to be $1$, then we have
\begin{align*}
P(y_i|X_i;\theta^*) \propto \exp \left\{ y_i \langle X_i,\theta^* \rangle - \frac{\langle X_i,\theta^* \rangle^2}{2} \right\} ~.
\end{align*}
Comparing it with the canonical form given earlier, the natural parameter is $\eta_i = \langle X_i,\theta^* \rangle$, log-partition function is $\varphi(\langle X_i,\theta^* \rangle) = \frac{\langle X_i,\theta^* \rangle^2}{2}$ and hence $E[y|X_i] = \varphi^{'}(\langle X_i,\theta^* \rangle ) = \langle X_i,\theta^* \rangle$. The noise $\omega_i = y_i - E(y_i|X_i)$ is Gaussian. Considering the negative log-likelihood, the GLM corresponding to the Gaussian distribution yields least squares regression~\cite{brow86}.

\item Bernoulli distribution: Assuming the conditional distribution of $y_i|X_i,\theta^*$ to have a Bernoulli distribution with conditional mean parameter $p_i$, which is a suitable function of $\langle X_i,\theta^*\rangle$, the likelihood of the observations is given by
\begin{align*}
P(y_i|p_i) &= p_i^{y_i} (1-p_i)^{(1-y_i)} = \exp(y_i \log p_i + (1-y_i) \log(1-p_i)) = \exp \left (y_i \log \left ( \frac{p_i}{1-p_i} \right ) + \log (1-p_i)  \right )
\end{align*}
Therefore, the natural parameter $\eta_i = \langle X_i,\theta^* \rangle = \log \left ( \frac{p_i}{1-p_i} \right )$ giving $p_i = \frac{\exp(\langle X_i,\theta^* \rangle)}{1 + \exp(\langle X_i,\theta^* \rangle)}$. It can be verified from the fact that the probability density function $P(y_i|p_i)$ adds to $1$ and the log-partition function evaluates to $\varphi(\eta_i) = \log(1 - p_i) = \log(1 + \exp(\langle X_i,\theta^* \rangle))$. The noise in the model corresponds to random draws from a Bernoulli distribution, and each element of $\omega$ is $\omega_i = p_i - y_i = \frac{\exp(\langle X_i,\theta^* \rangle)}{1 + \exp(\langle X_i,\theta^* \rangle)} - y_i$, which is bounded and hence sub-Gaussian. Considering the negative log-likelihood, the GLM corresponding to the Bernoulli distribution yields logistic regression~\cite{brow86}.

\item Poisson distribution: Assuming the conditional distribution of $y_i|X_i,\theta^*$ to have a Poisson distribution with conditional mean parameter $\lambda_i$, which is a suitable function of $\langle X_i,\theta^*\rangle$, the likelihood of the observations is given by
\begin{align*}
P(y_i|\lambda_i) &= \frac{\lambda_i^{y_i} \exp(-\lambda_i)}{y_i!}
\propto \exp \{ \log (\lambda_i^{y_i} \exp(-\lambda_i)) \} = \exp \{ y_i \log \lambda_i - \lambda_i \}~,
\end{align*}
where the $1/y_i!$ term constitutes the base measure for the distribution.
Based on the form, the natural parameter $\eta_i = \langle X_i,\theta^* \rangle = \log \lambda_i$ giving $\lambda_i = \exp(\eta_i) = \exp(\langle X_i,\theta^* \rangle)$. Also it can be verified that the log-partition function $\varphi(\eta_i) = \exp(\eta_i) = \lambda_i$. Each element of $\omega$ is $\omega_i = \lambda_i - y_i = \exp(\langle X_i,\theta^* \rangle) - y_i$.
Considering the negative log-likelihood, the GLM corresponding to the Poisson distribution yields Poisson regression~\cite{brow86}.

\end{enumerate}}

As discussed in Section~\ref{sec:glm}, if the design matrix $X$ and the noise is assumed to be sub-Gaussian, then the regularization parameter $\lambda_n$ needs to be $O(\frac{\Omega_R}{\sqrt{n}})$, following the analysis and results in Section~\ref{sec:lambda}. In the rest of this section, we focus on proving the GLMs with sub-Gaussian designs satisfy the RSC condition with sample complexity depending on the width of the spherical cap corresponding to the error set, as discussed in Section~\ref{sec:error}.

\subsection{RSC condition for GLMs}








For any convex loss function to satisfy the RSC condition on any $A \subseteq S^{p-1}$, the following inequality
\beq
\delta \cL(\theta^{*}, u;Z^{n}) = \cL(\theta^{*} + u; Z^{n}) - \cL(\theta^{*};Z^{n}) - \langle \nabla \cL(\theta^{*};Z^{n}), u \rangle \geq \kappa \|u\|_{2}^{2}
\eeq
needs to hold $\forall u \in A_{r}$. For the general formulation of GLM discussed earlier, we have
\begin{align*}
\delta \cL(\theta^{*}, u; Z^{n}) &= - \langle \theta^{*} + u, \frac{1}{n} \sum_{i=1}^{n}y_{i}X_{i} \rangle + \frac{1}{n} \sum_{i=1}^{n} \varphi(\langle \theta^{*} + u, X_{i} \rangle) + \langle \theta^{*}, \frac{1}{n} \sum_{i=1}^{n}y_{i}X_{i} \rangle \\
& \phantom{=} - \frac{1}{n} \sum_{i=1}^{n} \varphi(\langle \theta^{*}, X_{i} \rangle) - \langle -\frac{1}{n} \sum_{i=1}^{n} y_{i}X_{i} + \frac{1}{n} \sum_{i=1}^{n}X_{i} \varphi^{'}(\langle \theta^{*},x_{i} \rangle), u \rangle~.
\end{align*}
Simplifying the expression and applying mean value theorem twice we get the following
\beq
\delta \cL(\theta^{*}, u; Z^{n}) = \frac{1}{n} \sum_{i=1}^{n} \varphi^{''}\left ( \langle \theta^{*}, X_{i} \rangle + \gamma_{i} \langle u, X_{i} \rangle \right ) \langle u,X_{i} \rangle ^{2}~,
\eeq
for suitable $\gamma_{i} \in [0, 1]$. The RSC condition for GLMs then needs to consider lower bounds for
\beq
\delta \cL(\theta^{*}, u; Z^{n}) = \frac{1}{n} \sum_{i=1}^{n} \varphi^{''}(\langle \theta^{*}, X_{i} \rangle + \gamma_{i} \langle u, X_{i} \rangle) \langle u,X_{i} \rangle ^{2}
\eeq
where $\gamma_{i} \in [0, 1]$. The second derivative of the log-partition function is always positive. Since the RSC condition relies on a non-trivial lower bound for the above quantity, the analysis will suitably consider a compact set where $\ell = \ell_{\varphi}(T) = \min_{|a| \leq 2T} \varphi^{''}(a)$ is bounded away from zero. The only assumption outside this compact set $\{a: |a| \leq 2T \}$ is that the second derivative is greater than 0. Further, we assume $\| \theta^* \|_2 \leq c_1$ for some constant $c_1$. With these assumptions
\beq
\delta \cL (\theta^{*},u;Z^{n}) \geq \frac{\ell}{n} \sum_{i=1}^{n} \langle X_{i},u \rangle^{2} \mathbb{I}[| \langle X_{i}, \theta^{*} \rangle| < T] \mathbb{I}[| \langle X_{i},u \rangle | < T]~.
\eeq
We give a characterization of the RSC condition for isotropic sub-Gaussian design matrices $X \in \R^{n \times p}$. We consider $u \in A \subseteq S^{p-1}$ so that $\|u\|_2 = 1$. Further, we assume $\| \theta^* \|_2 \leq c_1$ for some constant $c_1$. Assuming $X$ has sub-Gaussian rows with $\vertiii{X_i}_{\varphi_2} \leq \kappa$,
$\langle X_{i}, \theta^{*} \rangle$ and $\langle X_{i}, u \rangle$ are sub-Gaussian random variables with sub-Gaussian norm at most $C\kappa$.

Let $\varepsilon_1$ and $\varepsilon_2$ denote the tail probability that $\langle X_i,u \rangle$ and $\langle X_i,\theta^* \rangle$ exceeds some constant $T$, i.e., $\varepsilon_1(T;u) = P\{ |\langle X_i, u \rangle| > T \} \leq e \cdot \exp( -c_{2}T^2/C^2 \kappa^2) = \bar{\varepsilon}_1$, and $\varepsilon_2(T;\theta^*) = P\{ | \langle X_i, \theta^* \rangle | > T \} \leq e \cdot \exp(- c_2 T^2/C^2 \kappa^2) = \bar{\varepsilon}_2$, where $\bar{\varepsilon}_1 = \bar{\varepsilon}_1(T,\kappa)$ and $\bar{\varepsilon}_2 = \bar{\varepsilon}_2(T;\kappa)$ are uniform upper bounds on the individual tail probabilities. The result we present below is in terms of the above defined constants $\ell = \ell_{\varphi}(T)$, $\bar{\varepsilon}_1 = \bar{\varepsilon}_1(T,\kappa)$ and $\bar{\varepsilon}_2 =  \bar{\varepsilon}_2(T,\kappa)$ for any suitably chosen $T$.

\textbf{Theorem \ref{thm:rscsubg}}
\textit{
Let $X \in \R^{n \times p}$ be a design matrix with independent isotropic sub-Gaussian rows such that $\vertiii{X_i}_{\varphi_2} \leq \kappa$. Then, for any set $A \subseteq S^{p-1}$ for suitable constants $\eta,c > 0$, with probability at least $1- 2 \exp \left(-\eta w^2(A) \right)$, we have
\beq
\inf_{u \in A} \partial \cL (\theta^{*};u,X) \geq \ell \underline{\rho}^2 \left (1 - c\kappa_1^2 \frac{w(A)}{\sqrt{n}} \right ) ~.
\eeq
where $\underline{\rho}^2 = \inf_{u \in A} \rho_u^2$, with $\rho_u^2 = E[\langle X_{i},u \rangle^{2} \mathbb{I}[| \langle X_{i}, \theta^{*} \rangle| < T] \mathbb{I}[| \langle X_{i},u \rangle | < T ]]$, and $\kappa_1 = \frac{\kappa}{(1-\bar{\varepsilon}_1-\bar{\varepsilon}_2)^2}$.
}
\proof For any fixed $T$, let $\bar{Z}_i = \bar{Z}_i^u = \langle X_i, u \rangle \I(|\langle X_i, u \rangle| \leq T) \I(|\langle X_i, \theta^{*} \rangle| \leq T)$. Then, the probability distribution over $\bar{Z}_i$ can be written as:\footnote{With abuse of notation, we treat the distribution over $\bar{Z}_i$ as discrete for ease of notation. A similar argument applies for the true continuous distribution, but more notation is needed.}
\beq
P(\bar{Z}_i = z) = \frac{P(\langle X_i,u \rangle = z)\I(|\langle X_i, u \rangle| \leq T)\I(|\langle X_i, \theta^{*} \rangle| \leq T)}{P(|\langle X_i, u \rangle| \leq T, |\langle X_i, \theta^{*} \rangle| \leq T)} \leq \frac{1}{1-\bar{\varepsilon}_1 - \bar{\varepsilon}_2}P(\langle X_i,u \rangle = z)~.
\eeq
As a result, $\vertiii{\bar{Z}_i}_{\psi_2} \leq \frac{\kappa}{1- \bar{\varepsilon}_1 - \bar{\varepsilon}_2} = \kappa_1$. Thus, $\bar{Z}_i = \bar{Z}_i^u$ is a sub-Gaussian random variable for any $u \in A$. Let $\rho_u^2 = E[(\bar{Z_i}^u)^2] > 0$.
Let $X_0$ be i.i.d. as the rows $X_i,i = 1,\hdots,n$. Let $A \subseteq S^{p-1}$ and consider the following class of functions: $F = \{ \frac{1}{\rho_u} \langle \cdot,u \rangle \I(|\langle \cdot, u \rangle| \leq T)\I(|\langle \cdot, \theta^{*} \rangle| \leq T)  : u \in A \}$. Then for any $f \in F$, $f(X_0) = \frac{1}{\rho_u}\langle X_0,u \rangle \I(|\langle X_0, u \rangle| \leq T)\I(|\langle X_0, \theta^{*} \rangle| \leq T)$ and, by construction, $F$ is a subset of the unit sphere, i.e., $F \subseteq S_{L_2}$. Further, $\sup_{f \in F} \vertiii{f}_{\psi_2} \leq \kappa_1/2$.

Next, we show that for the current setting, the $\gamma_2$-functional can be upper bounded by $w(A)$, the Gaussian width of $A$. Since the process is sub-Gaussian with $\varphi_2$-norm bounded by $\kappa_1$, we have
\beq
\gamma_2(F \cap S_{L_2}, \vertiii{\cdot}_{\psi_2}) \leq \kappa_1 \gamma_2(F \cap S_{L_2},\vertiii{\cdot}_{L_2}) \leq \kappa_1 c_4 w(A)~,
\eeq
where the last inequality follows from generic chaining, in particular \cite[Theorem 2.1.1]{tala05}, for an absolute constant $c_4 > 0$.

In the context of Theorem~\ref{thm:subGREmain}, we choose
\beq
\theta = c_1 c_4 \kappa_1^2 \frac{w(A)}{\sqrt{n}} \geq c_1 \kappa_1 \frac{\gamma_2(F \cap S_{L_2},\vertiii{\cdot}_{\varphi_2})}{\sqrt{n}} ,
\eeq
so that the condition on $\theta$ is satisfied. With this choice of $\theta$, we have
\beq
\theta^2 n /\kappa_1^4 = c_1^2 c_4^2 w^2(A) ~.
\eeq
Then, from Theorem~\ref{thm:subGREmain}, it follows that with probability at least $1 - \exp(-\eta w^2(A))$, we have
\beq
\sup_{u \in A} \left | \frac{1}{\rho_u n} \sum_{i=1}^{n} \langle X_i,u \rangle^2 \I(|\langle X_0, u \rangle| \leq T)\I(|\langle X_0, \theta^{*} \rangle| \leq T) - 1 \right | \leq c \kappa_1^2\frac{w(A)}{\sqrt{n}}
\eeq
where $\eta = c_2 c_1^2 c_4^2$ and $c = c_1 c_2$ are absolute constants. Thus, with probability at least $1 - \exp(-\eta w^2(A))$,
\beq
\inf_{u \in A} \frac{1}{n} \sum_{i=1}^{n} \langle X_i,u \rangle^2 \I(|\langle X_0, u \rangle| \leq T)\I(|\langle X_0, \theta^{*} \rangle| \leq T) \geq \inf_{u \in A} ~\rho_u^2 \left( 1 - c  \kappa_1^2 \frac{w(A)}{\sqrt{n}} \right)~.
\eeq
Denoting $\underline{\rho}^2 = \inf_{u \in A} \rho_u^2$, with probability at least $1 - \exp(-\eta w^2(A))$, we have
\beq
\inf_{u \in A} \partial \cL (\theta^{*};u,X) \geq \inf_{u \in A} ~\frac{\ell}{n} \sum_{i=1}^{n} \langle X_{i},u \rangle^{2} \mathbb{I}[| \langle X_{i}, \theta^{*} \rangle| < T] \mathbb{I}[| \langle X_{i},u \rangle | < T]
 \geq \ell \underline{\rho}^2 \left ( 1 - c\kappa_1^2 \frac{w(A)}{\sqrt{n}} \right)	~.
\eeq	
That completes the proof. \qed

{\bf Acknowledgements:} We thank the reviewers of the conference version~\cite{bcfs14} for helpful comments and suggestions on related work. We thank Sergey Bobkov, Snigdhansu Chatterjee, and Pradeep Ravikumar for helpful discussions related to the paper. The research was supported by NSF grants IIS-1447566, IIS-1422557, CCF-1451986, CNS-1314560, IIS-0953274, IIS-1029711,  and by NASA grant NNX12AQ39A.

\bibliographystyle{plain}
\bibliography{supp_norm_ref}

\begin{thebibliography}{10}

\bibitem{almt14}
D.~Amelunxen, M.~Lotz, M.~B. McCoy, and J.~A. Tropp.
\newblock {Living on the edge: Phase transitions in convex programs with random
  data}.
\newblock {\em Inform. Inference}, 3(3):224--294, 2014.

\bibitem{bcfs14}
A.~Banerjee, S.~Chen, F.~Fazayeli, and V.~Sivakumar.
\newblock Estimation with norm regularization.
\newblock In {\em Advances in Neural Information Processing Systems (NIPS)},
  2014.

\bibitem{bmdg05}
A~Banerjee, S~Merugu, I~Dhillon, and J~Ghosh.
\newblock {Clustering with {B}regman Divergences}.
\newblock {\em Journal of Machine Learning Research}, 6:1705--1749, 2005.

\bibitem{barn78}
O~Barndorff-Nielsen.
\newblock {\em {Information and Exponential Families in Statistical Theory}}.
\newblock John Wiley and Sons, 1978.

\bibitem{bame02}
P.~L. Bartlett and S.~Mendelson.
\newblock {Rademacher and Gaussian complexities: Risk bounds and structural
  results}.
\newblock {\em Journal of Machine Learning Research}, 3:463--482, 2002.

\bibitem{birt09}
P.~J. Bickel, Y.~Ritov, and A.~B. Tsybakov.
\newblock {Simultaneous analysis of Lasso and Dantzig selector}.
\newblock {\em The Annals of Statistics}, 37(4):1705--1732, 2009.

\bibitem{bolm13}
S.~Boucheron, G.~Lugosi, and P.~Massart.
\newblock {\em Concentration {Inequalities}: {A} {Nonasymptotic} {Theory} of
  {Independence}}.
\newblock Oxford University Press, 2013.

\bibitem{brow86}
L~Brown.
\newblock {\em {Fundamentals of Statistical Exponential Families}}.
\newblock Institute of Mathematical Statistics, 1986.

\bibitem{buva11}
P.~Buhlmann and S.~van~de Geer.
\newblock {\em {Statistics for High Dimensional Data: Methods, Theory and
  Applications}}.
\newblock Springer Series in Statistics. Springer, 2011.

\bibitem{cand08}
E.~Candes.
\newblock {The restricted isometry property and its implications for compressed
  sensing}.
\newblock {\em Comptes Rendus Mathematique}, 346(9-10):589--592, 2008.

\bibitem{cata07}
E.~Candes and T~Tao.
\newblock {The {D}antzig selector: statistical estimation when {$p$} is much
  larger than {$n$}}.
\newblock {\em The Annals of Statistics}, 35(6):2313--2351, 2007.

\bibitem{cart06}
E.~J. Candes, J.~Romberg, and T.~Tao.
\newblock {Robust uncertainty principles: Exact signal reconstruction from
  highly incomplete frequency information}.
\newblock {\em IEEE Transactions on Information Theory}, 52:489--509, 2006.

\bibitem{cata05}
E.~J. Candes and T.~Tao.
\newblock {Decoding by linear programming}.
\newblock {\em IEEE Transactions on Information Theory}, 51:4203--4215, 2005.

\bibitem{crpw12}
V.~Chandrasekaran, B.~Recht, P.~A. Parrilo, and A.~S. Willsky.
\newblock The convex geometry of linear inverse problems.
\newblock {\em Foundations of Computational Mathematics}, 12(6):805--849, 2012.

\bibitem{chcb14}
S.~Chatterjee, S.~Chen, and A.~Banerjee.
\newblock Generalized dantzig selector: Application to the k-support norm.
\newblock In {\em Advances in Neural Information Processing Systems (NIPS)},
  2014.

\bibitem{chba15}
S.~Chen and A.~Banerjee.
\newblock Structured estimation with atomic norms: General bounds and
  applications.
\newblock In {\em Advances in Neural Information Processing Systems (NIPS)},
  2015.

\bibitem{dagu03}
Sanjoy Dasgupta and Anupam Gupta.
\newblock An elementary proof of a theorem of johnson and lindenstrauss.
\newblock {\em Random Struct. Algorithms}, 22(1):60--65, 2003.

\bibitem{fino14}
M.~A.~T. Figueiredo and R.~D. Nowak.
\newblock Sparse estimation with strongly correlated variables using ordered
  weighted l1 regularization.
\newblock {\em arXiv:1409.4005}, 2014.

\bibitem{gord85}
Y.~Gordon.
\newblock Some inequalities for gaussian processes and applications.
\newblock {\em Israel Journal of Mathematics}, 50(4):265--289, 1985.

\bibitem{gord88}
Y.~Gordon.
\newblock On milman's inequality and random subspaces which escape through a
  mesh in $r^n$.
\newblock In {\em Geometric Aspects of Functional Analysis}, volume 1317 of
  {\em Lecture Notes in Mathematics}, pages 84--106. Springer, 1988.

\bibitem{klme05}
B.~Klartag and S.~Mendelson.
\newblock Empirical processes and random projections.
\newblock {\em Journal of Functional Analysis}, 225(1):229--245, 2005.

\bibitem{kome13}
V.~Koltchinskii and S.~Mendelson.
\newblock {Bounding the smallest singular value of a random matrix without
  concentration}.
\newblock {\em arXiv:1312.3580}, 2013.

\bibitem{leme14}
G.~Lecu\'{e} and S.~Mendelson.
\newblock {Sparse recovery under weak moment assumptions}.
\newblock {\em arXiv:1401.2188}, 2014.

\bibitem{ledo13}
M.~Ledoux and M.~Talagrand.
\newblock {\em {Probability in Banach Spaces: Isoperimetry and Processes}}.
\newblock Springer, 2013.

\bibitem{mapr14}
A.~Maurer, M.~Pontil, and B.~Romera-Paredes.
\newblock {An Inequality with Applications to Structured Sparsity and Multitask
  Dictionary Learning}.
\newblock In {\em Conference on Learning Theory (COLT)}, 2014.

\bibitem{meyu09}
N.~Meinshausen and B.~Yu.
\newblock {Lasso-type recovery of sparse representations for high-dimensional
  data}.
\newblock {\em The Annals of Statistics}, 37(1):246---270, 2009.

\bibitem{mend14}
S.~Mendelson.
\newblock {Learning Without Concentration}.
\newblock In {\em Journal of the ACM}, 2015.

\bibitem{mept07}
S.~Mendelson, A.~Pajor, and N.~Tomczak-Jaegermann.
\newblock Reconstruction and sub{G}aussian operators in asymptotic geometric
  analysis.
\newblock {\em Geometric and Functional Analysis}, 17:1248--1282, 2007.

\bibitem{nrwy12}
S.~Negahban, P.~Ravikumar, M.~J. Wainwright, and B.~Yu.
\newblock {A unified framework for the analysis of regularized $M$-estimators}.
\newblock {\em Statistical Science}, 27(4):538--557, 2012.

\bibitem{oliv13}
R.~I. Oliveira.
\newblock {The lower tail of random quadratic forms, with applications to
  ordinary least squares and restricted eigenvalue properties}.
\newblock {\em arXiv:1312.2903}, 2013.

\bibitem{oyth13}
S.~Oymak, C.~Thrampoulidis, and B.~Hassibi.
\newblock {The Squared-Error of Generalized Lasso: A Precise Analysis}.
\newblock {\em arXiv:1311.0830}, 2013.

\bibitem{plve13}
Y.~Plan and R.~Vershynin.
\newblock {Robust 1-bit compressed sensing and sparse logistic regression: A
  convex programming approach}.
\newblock {\em IEEE Transactions on Information Theory}, 59(1):482--494, 2013.

\bibitem{rawy10}
G.~Raskutti, M.~J. Wainwright, and B.~Yu.
\newblock {Restricted Eigenvalue Properties for Correlated Gaussian Designs}.
\newblock {\em Journal of Machine Learning Research}, 11:2241--2259, 2010.

\bibitem{ruzh13}
Z.~Rudelson and S.~Zhou.
\newblock Reconstruction from anisotropic random measurements.
\newblock {\em IEEE Transactions on Information Theory}, 59(6):3434--3447,
  2013.

\bibitem{rudi76}
Walter Rudin.
\newblock {\em {Principles of Mathematical Analysis}}.
\newblock International Series in Pure \& Applied Mathematics. McGraw-Hill, 3rd
  edition, 1976.

\bibitem{sibr15}
V.~Sivakumar, A.~Banerjee, and P.~Ravikumar.
\newblock Beyond sub-gaussian measurements: High-dimensional structured
  estimation with sub-exponential designs.
\newblock In {\em Advances in Neural Information Processing Systems (NIPS)},
  2015.

\bibitem{tala05}
M.~Talagrand.
\newblock {\em The Generic Chaining}.
\newblock Springer, 2005.

\bibitem{tala14}
M.~Talagrand.
\newblock {\em Upper and Lower Bounds for Stochastic Processes}.
\newblock Springer, 2014.

\bibitem{tibs96}
R.~Tibshirani.
\newblock {Regression shrinkage and selection via the Lasso}.
\newblock {\em Journal of the Royal Statistical Society, Series B},
  58(1):267--288, 1996.

\bibitem{trop14}
J.~A. Tropp.
\newblock Convex recovery of a structured signal from independent random linear
  measurements.
\newblock In {\em Sampling Theory, a Renaissance}. (To Appear), 2015.

\bibitem{vers12}
R.~Vershynin.
\newblock Introduction to the non-asymptotic analysis of random matrices.
\newblock In Y.~Eldar and G.~Kutyniok, editors, {\em Compressed Sensing},
  chapter~5, pages 210--268. Cambridge University Press, 2012.

\bibitem{vers14}
R.~Vershynin.
\newblock Estimation in high dimensions: {A} geometric perspective.
\newblock {\em Submitted}, 2014.

\bibitem{wain09}
M.~J. Wainwright.
\newblock {Sharp thresholds for noisy and high-dimensional recovery of sparsity
  using $\ell_1$-constrained quadratic programming(Lasso)}.
\newblock {\em IEEE Transactions on Information Theory}, 55:2183--2202, 2009.

\bibitem{wajo08}
M~J Wainwright and M~I Jordan.
\newblock {Graphical models, exponential families, and variational inference}.
\newblock {\em Foundations and Trends in Machine Learning}, 1(1-2):1--305,
  2008.

\bibitem{zhyu06}
P.~Zhao and B.~Yu.
\newblock {On model selection consistency of Lasso}.
\newblock {\em Journal of Machine Learning Research}, 7:2541--2567, November
  2006.

\bibitem{zhou09}
S.~Zhou.
\newblock {Restricted eigenvalue conditions on subgaussian random matrices}.
\newblock Technical report, Department of Mathematics, ETH Zurich, December
  2009.

\end{thebibliography}

\end{document}